\documentclass[10pt,final,journal]{IEEEtran}
\IEEEoverridecommandlockouts                              % This command is only 
\usepackage[left=0.75in,right=0.75in,top=0.75in,bottom=0.75in]{geometry}
\usepackage{amsfonts}
\usepackage{algorithmic}
\usepackage{graphicx}
\usepackage{textcomp}
\usepackage{xcolor}
\usepackage[hidelinks]{hyperref}
\usepackage{rotating}
\usepackage{multicol}
\usepackage{multirow}
\usepackage{graphics} % for pdf, bitmapped graphics files
\usepackage{epsfig} % for postscript graphics files
\usepackage{times} % assumes new font selection scheme installed
\usepackage{amsmath} % assumes amsmath package installed
\usepackage{amssymb}  % assumes amsmath package installed
\usepackage{cite}
\usepackage{url}
\usepackage{dblfloatfix}    % To enable figures at the bottom of page
\usepackage{easyReview}
\usepackage{siunitx}

\usepackage[pscoord]{eso-pic}
\newcommand{\placetextbox}[3]{
\setbox0=\hbox{#3}
\AddToShipoutPictureFG{ \put(\LenToUnit{#1\paperwidth},\LenToUnit{#2\paperheight}){\vtop{{\null}\makebox[0pt][c]{#3}}}}
}
\placetextbox{.5}{0.035}{~\copyright 2021 IEEE. Personal use of this material is permitted. Permission from IEEE must be obtained for all other uses.}

\newcommand{\orcidicon}{\includegraphics[width=0.32cm]{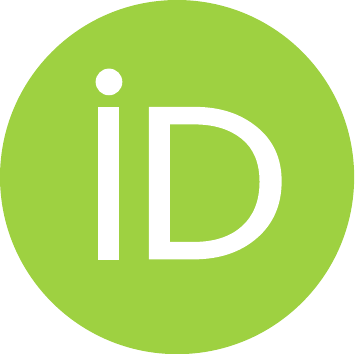}}
\expandafter\xdef\csname orcidCK\endcsname{\noexpand\href{https://orcid.org/\csname orcidauthorCK\endcsname}{\noexpand\orcidicon}}
\expandafter\xdef\csname orcidBM\endcsname{\noexpand\href{https://orcid.org/\csname orcidauthorBM\endcsname}{\noexpand\orcidicon}}
\expandafter\xdef\csname orcidCT\endcsname{\noexpand\href{https://orcid.org/\csname orcidauthorCT\endcsname}{\noexpand\orcidicon}}
\expandafter\xdef\csname orcidJG\endcsname{\noexpand\href{https://orcid.org/\csname orcidauthorJG\endcsname}{\noexpand\orcidicon}}
\expandafter\xdef\csname orcidGP\endcsname{\noexpand\href{https://orcid.org/\csname orcidauthorGP\endcsname}{\noexpand\orcidicon}}
\expandafter\xdef\csname orcidJB\endcsname{\noexpand\href{https://orcid.org/\csname orcidauthorJB\endcsname}{\noexpand\orcidicon}}
\expandafter\xdef\csname
orcidSD\endcsname{\noexpand\href{https://orcid.org/\csname orcidauthorSD\endcsname}{\noexpand\orcidicon}}
\expandafter\xdef\csname
orcidYG\endcsname{\noexpand\href{https://orcid.org/\csname orcidauthorYG\endcsname}{\noexpand\orcidicon}}
\expandafter\xdef\csname
orcidJS\endcsname{\noexpand\href{https://orcid.org/\csname orcidauthorJS\endcsname}{\noexpand\orcidicon}}
\expandafter\xdef\csname
orcidDR\endcsname{\noexpand\href{https://orcid.org/\csname orcidauthorDR\endcsname}{\noexpand\orcidicon}}

\renewcommand{\IEEEtitletopspaceextra}{0.25in\relax}

\usepackage[flushleft]{threeparttable}
\newcommand{\cmmnt}[1]{\ignorespaces}
\title{\LARGE \bf
NASA Space Robotics Challenge 2 Qualification Round: An Approach to Autonomous Lunar Rover Operations
}
\author{Cagri Kilic\IEEEauthorrefmark{1}\orcidCK{}, Bernardo Martinez R. Jr.\IEEEauthorrefmark{1}\orcidBM{}, Christopher A. Tatsch\IEEEauthorrefmark{1}\orcidCT{}, Jared Beard\orcidJB{}, Jared Strader\orcidJS{}, \\ Shounak Das\orcidSD{}, Derek Ross\orcidDR{}, Yu Gu\orcidYG{}, Guilherme A. S. Pereira\orcidGP{}, and Jason N. Gross\IEEEauthorrefmark{2}\orcidJG{} % <-this % stops a space
 \thanks{This work was supported in part by Statler College of Engineering and  Mineral Resources of West Virginia University.}
 \thanks{Authors are with the Department of Mechanical and Aerospace Engineering,  
West Virginia University, Morgantown, USA. 
}
\thanks{
\IEEEauthorrefmark{1}{ These authors contributed equally to this paper. } 
}
\thanks{
\IEEEauthorrefmark{2}{ Corresponding Author: Jason N. Gross ( Jason.Gross@mail.wvu.edu )} 
}
}

\begin{document}

\markboth{IEEE Aerospace and Electronic Systems Magazine. Preprint Version. Accepted September, 2021}
{NASA Space Robotics Challenge 2 Qualification Round: An Approach to Autonomous Lunar Rover Operations} 

\maketitle

\begin{abstract}
Plans for establishing a long-term human presence on the Moon will require substantial increases in robot autonomy and multi-robot coordination to support establishing a lunar outpost. To achieve these objectives, algorithm design choices for the software developments need to be tested and validated for expected scenarios such as autonomous in-situ resource utilization (ISRU), localization in challenging environments, and multi-robot coordination. However, real-world experiments are extremely challenging and limited for extraterrestrial environment. Also, realistic simulation demonstrations in these environments are still rare and demanded for initial algorithm testing capabilities. To help some of these needs, the NASA Centennial Challenges program established the Space Robotics Challenge Phase 2 (SRC2) which consist of virtual robotic systems in a realistic lunar simulation environment, where a group of mobile robots were tasked with reporting volatile locations within a global map, excavating and transporting these resources, and detecting and localizing a target of interest. The main goal of this article is to share our team's experiences on the design trade-offs to perform autonomous robotic operations in a virtual lunar environment and to share strategies to complete the mission requirements posed by NASA SRC2 competition during the qualification round. Of the 114 teams that registered for participation in the NASA SRC2, team Mountaineers finished as one of only six teams to receive the top qualification round prize.

\end{abstract}

\begin{IEEEkeywords}
Autonomy, In-Situ Resource Utilization, Lunar Planetary Rovers, Robotics Competitions
\end{IEEEkeywords}

\section{Introduction}
\label{sec:introduction}
In-situ resource utilization (ISRU) in extraterrestrial soil will allow continuous and affordable human discovery of many deep-space destinations~\cite{colaprete2017resource}. Essential resources like oxygen and water on the Moon can be used as both vital consumables for humans and building materials of rocket fuel. Moreover, new observations of the Moon missions (both orbital and surface) have provided evidence of a lunar water formation that is more complex and rich than previously believed~\cite{colaprete2017resource}. While the proof of the existence of lunar resources is increasing, the distribution of these resources is not well-known~\cite{sanders2012progress}. 

Aiming to find and use water and other essential resources, to learn how to live and operate on the surface of another celestial body, and to learn more about our own planet and the Moon, NASA has launched the Artemis program~\cite{artemisplan2020,artemis2020Smith}. This program, that will take the first woman and the next man to the lunar surface, is currently underway to meet the agency's exploration goals. The Artemis Plan will culminate in the foundations for a sustained long-term presence on the Moon and prepare for future presence in other planets, and, more specifically, Mars~\cite{mendell2005meditations}. 

Consonant with this strategy, NASA is planning a series of progressive robotic missions to the lunar surface. According to the Artemis plan~\cite{artemisplan2020}, first, the lunar soil will be extensively explored by scout robots to confirm the information collected using orbital missions.
\begin{figure}[tb!]
    \centering
    \includegraphics[trim={8cm 11cm 8cm 0}, width=\linewidth, clip]{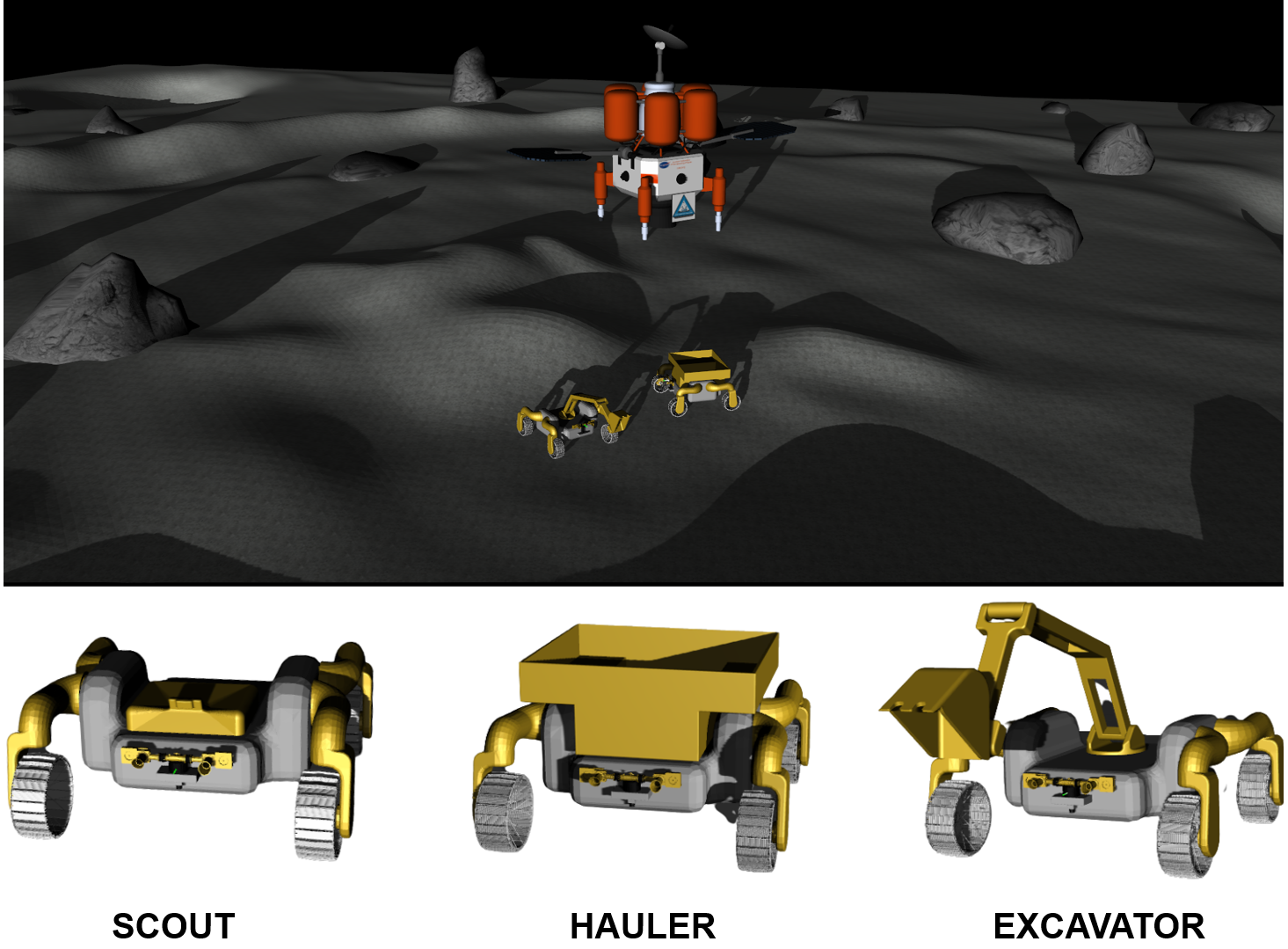}
    \caption{An illustration of the virtual lunar environment and two of the rovers provided by the competition for the SRC2 qualification round. The lunar environment consists of hills, slopes, rocks, craters, resources, and a processing plant. }
    \label{fig:environment}
\end{figure}
Later, rovers and landers will test technologies developed to amplify the capabilities on the Moon, such as robotic mining and energy storage systems. These robots will be deployed in the lunar South Pole and establish the Artemis Base Camp~\cite{artemis2020Smith}. This region was chosen because it has access to both permanently shadowed regions, where necessary resources are believed to be present, and regions that are exposed to sunlight for extended lengths of time during the year, guaranteeing energy for powering all these robotic systems~\cite{artemisplan2020}.  

Concurrently, NASA has also launched a Lunar Surface Innovation Initiative that aims to advance the following capabilities~\cite{LSI2020}: 1) Lunar in-situ resource utilization, 2) establishment of sustainable power during the lunar day/night cycles, 3) building machinery resistant to extreme environmental conditions, 4) lunar dust mitigation, 5) execution of surface excavation, manufacturing, and construction duties, and 6) extreme access including navigation and exploration of the lunar surface and subsurface.
\begin{figure}[tb!]
    \centering
    \includegraphics[width=\linewidth]{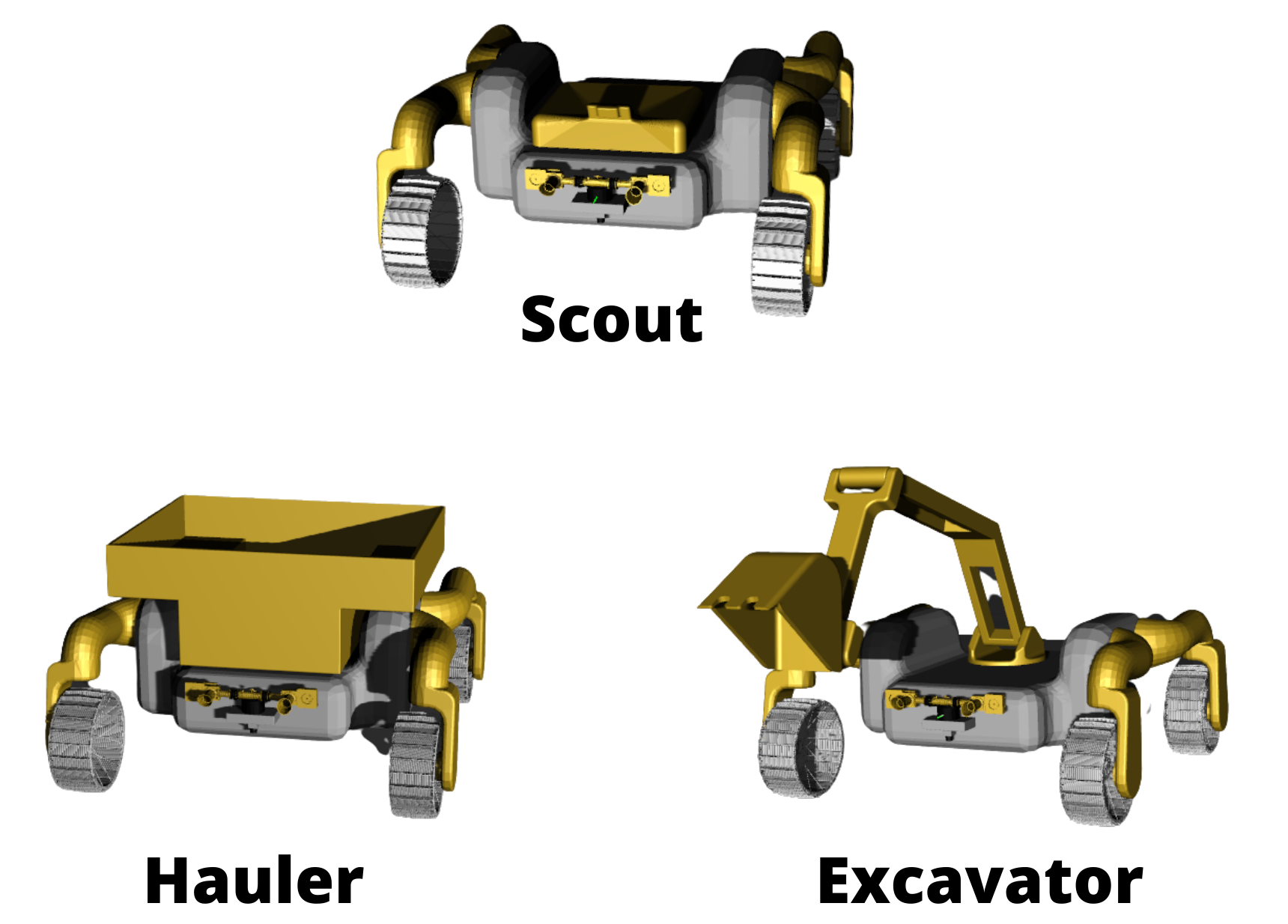}
    \caption{Depictions of the provided rovers in the competition qualification round. These rovers were expansions of a medium sized four wheeled robot, called as Base Rover. Each rover was designed by the competition for a specific goal in the qualification round tasks. The Scout rover was used for resource exploration and localization (tasks 1 and 3). The Hauler and the Excavator rovers were collaboratively used in resource collection and excavation, respectively (task 2).  See Section~\ref{sec:overview} for detailed explanation of the tasks.}
    \label{fig:rovers}
\end{figure}
Along with this initiative, the NASA Centennial Challenges Program (CCP) has sponsored the SRC2~\cite{SRC2Rules2020}, which is a prize competition that was launched to engage the public in developing solutions to allow heterogeneous multi-robot teams to autonomously complete tasks envisioned for ISRU, extreme access, and excavation in a virtual lunar environment. The challenge consists of a qualification round and a competition round. Both rounds will require fully autonomous operations that are robust enough to handle a randomized environment upon each trial. This article's main goal is to share the experiences and insights during our participation in the SRC2 challenge qualification round with the community.  A pre-competition team report can be found in~\cite{teammountaineers2020}. 

Through this article, the following contributions are presented: 1) a thorough discussion of the problems encountered in the SRC2 challenge, 2) specific capabilities implemented by our team to support autonomous resource localization, resource excavation, and object detection tasks, 3) a potential solution to future cooperative robotic lunar exploration. Out of 114 teams that registered to participate in the NASA SRC2 qualification round, our team was one of only six teams that qualified while scoring enough points to earn the top prize. A video attachment of our team's qualification round submission is provided for the interested readers\footnote{\url{https://youtu.be/S4-EzKoEqSk}}. 

The rest of the paper, is organized as follows. Section~\ref{sec:overview} summarizes the qualification round specifications. Section~\ref{sec:system_design} describes the main capabilities. The developed task strategies are summarized in Section~\ref{sec:strategies}. Section~\ref{sec:lesson_learned} explores the technical challenges faced during the qualification round and the goals for the final round. Finally, conclusions are presented in Section~\ref{sec:conclusion}.

\section{Overview of the Challenge} 
\label{sec:overview}
\begin{figure*}[b!]
    \centering
    \includegraphics[width=0.945\linewidth]{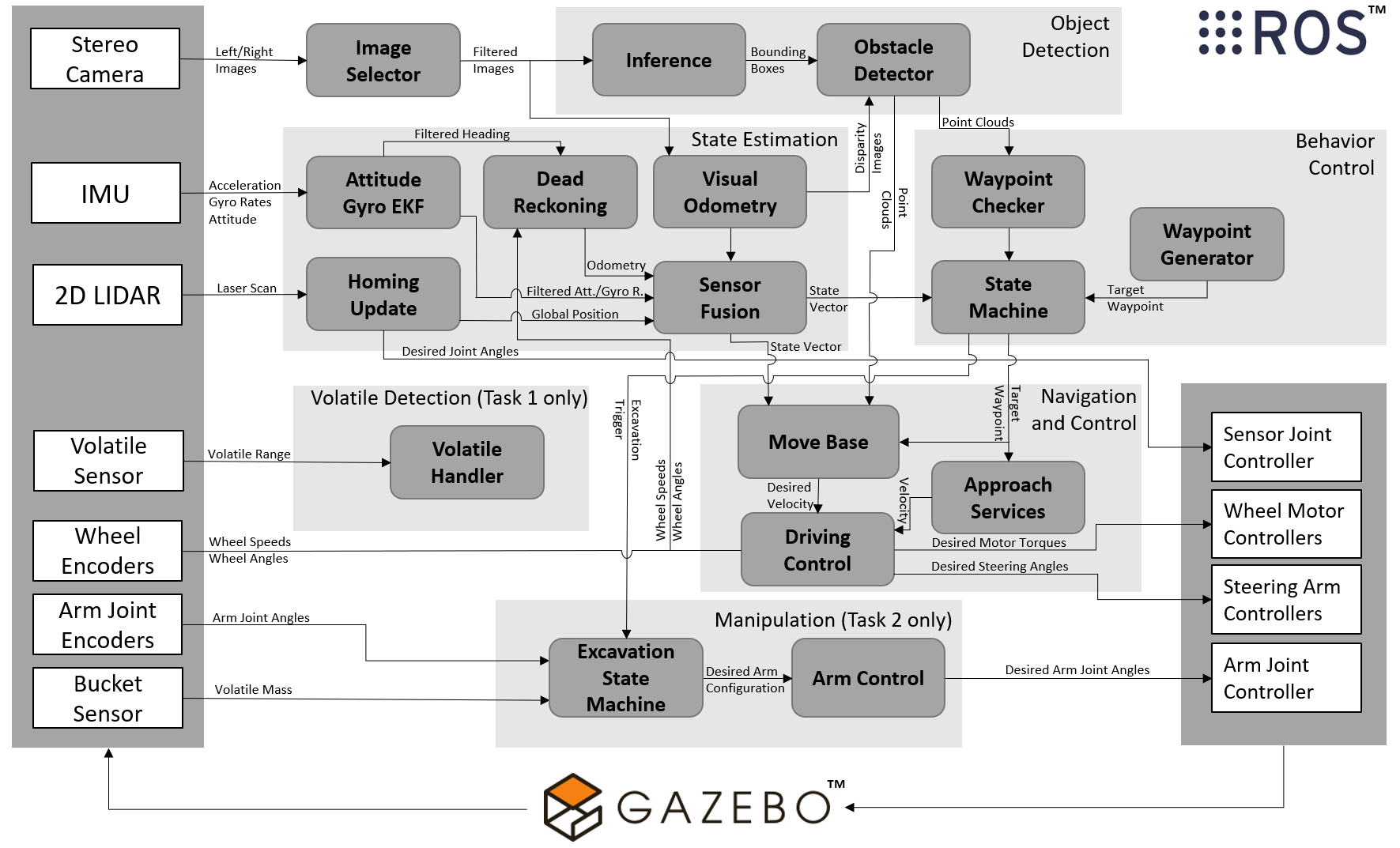}
    \caption{The architecture of the designed robotic system. Interactions of the robots with the environment are simulated using Gazebo. The white blocks represent the interfaces with the simulator. Sensors are represented on the left block, and robot actuators are represented on the right block. The grey blocks represent ROS software developed by the team. The arrows connecting the the blocks show the data being streamed between them.}
    \label{fig:architecture}
\end{figure*}

The SRC2 qualification round consisted of three tasks to be completed by virtual robotic systems in a simulated environment provided by the competition organizers. These tasks motivated the development of reliable software to advance the surface mining capabilities of fully autonomous robots on the lunar surface. Brief descriptions for the tasks are provided below, and the full descriptions can be found in the official rule document~\cite{SRC2Rules2020}.

\begin{itemize}
    \item \textbf{Task 1:}
The goal of task 1 was to explore, detect, and identify lunar resources that consisted of 28 different volatiles randomly distributed within a simulated lunar environment. The Scout rover (see Fig.~\ref{fig:rovers}) was used for this task, which has a volatile sensor capable of resource detection and identification when the rover is within 2\,\si{m} of a volatile. To locate any resource successfully, the locations of the sensed volatiles must be reported to the competition scoring system with an accuracy of 2\,\si{m} of the actual resource location.

    \item \textbf{Task 2:}
The goal of task 2 was to excavate resources at varying depths below the surface. The competition provided an Excavator and a Hauler for performing this task (see Fig.~\ref{fig:rovers}). In contrast to task 1, the locations of the resources were provided \textit{a priori} and the problem consisted of reaching the resource locations, digging the volatiles with the Excavator, and dropping them in the Hauler's bin. Resources were considered collected and awarded points if they were dispensed entirely in the Hauler's bin.

\item \textbf{Task 3:}
The goal of task 3 was to demonstrate the capability of object localization and robot alignment with the base station. In this task, an \textit{a priori} known object, modeled as a CubeSat, was randomly placed in the simulation world. The object was stationary and above the surface. The rover was required to report the location of the CubeSat within the accuracy of 5\,\si{m}. The CubeSat was physically unreachable by the rovers and placed at a random height between 5\,\si{m} and 25\,\si{m} above the surface of the virtual lunar environment. Additionally, after reporting the object position, the rover should find the processing plant, approach, and align itself with a specific marker on the station.
\end{itemize}

As mentioned before, there were three lunar rovers provided by the competition, as shown in Fig.~\ref{fig:rovers}. 
These rovers were all expansions of a ``base rover'', a medium-sized four-wheeled robot with individual control of the wheel steering angles and motor torques. The torque and velocity of the wheels were constrained so that the rover would only cruise at a maximum speed of 1.5~\si{m/s}, and the steering angle was constrained to $\pm$ 90~\si{deg}, allowing a great range of driving possibilities. The competition provided a tuned proportional-integral-derivative (PID) controller for controlling the steering angles and motor torques of each wheel. The base rover was equipped with sensors, including a planar LiDAR, a stereo camera, and an inertial measurement unit (IMU) to support localization and perception. The LiDAR and cameras were actuated and could be tilted up and down.

Each specialized rover was adapted for its task as described below:

\begin{itemize}
    \item \textbf{Scout}: designed for task 1 and task 3, it was equipped with a sensor that can detect and identify the volatiles in the environment.
    \item \textbf{Hauler}: designed for task 2, it was designed to transport collected resources back to the processing plant.
    \item \textbf{Excavator}: designed for task 2, it was designed to excavate resources below the lunar surface with a four degrees-of-freedom (DoF) manipulator with a bucket end-effector.
\end{itemize}

The SRC2 lunar and robot models utilized the simulation environment Gazebo\texttrademark ~\cite{GazeboTutorials}, which offers an interface with the Robot Operating System (ROS\texttrademark)\cite{ROSdocumentation}. ROS is a framework that facilitates the development of robotics software through hardware abstractions and interfaces, package management, and inter-program communication. Central to ROS is the approach it takes to facilitate the information flow between programs, referred to as nodes. Any node can read/write (publish/subscribe) to ROS topics sending messages and allowing the information to be accessible by several nodes simultaneously. A ROS service carries out a task and provides information about this task to a client node which requests the task be carried out. 

In the qualification round, the competitors were expected to overcome several hardware constraints and technical challenges similar to a planetary exploration mission as follows:     
\begin{enumerate}
    \item Having no GPS or similar satellite based lunar system for localization.
    \item Having no communication with the base station (e.g. beacon signal) requiring full autonomy for the rovers.
    \item Using coupled and limited range sensor package (LiDAR and stereo camera).
    \item Using single stereo camera in a low feature, dark environment with some permanently shadowed areas which impair visual odometry (VO) performance.
    \item Dealing with steep slopes in the terrain that create significant slip and prevent the rovers from climbing in a crater and stop on hills.
    \item Dealing with limited detection of volatiles to short range distances, which could only be performed by a specialized rover.
    \item Dealing with randomly distributed obstacles, volatile locations, initial rover and processing facility poses, and CubeSat (i.e, for task 3) for each simulation seed. 
    \item Working with time limitation (45 minutes).
\end{enumerate}

To overcome these constraints and challenges, we identified and designed the subsystems and capabilities as detailed in Section~\ref{sec:system_design} and implemented them in our task strategies as explained in Section~\ref{sec:strategies}.

\section{Systems Design}
\label{sec:system_design}

This section provides detailed information regarding the main capabilities of the system developed to overcome the challenges listed in the previous section. A complete system architecture with provided sensors, actuators, and corresponding inputs and outputs is shown in Fig.~\ref{fig:architecture}. The interactions between the main robot capabilities shown in this figure are: State Estimation, Navigation and Control, Object Detection, and Manipulation. The details of each of these subsystems are presented in the next subsections. Behavior control and other details of Volatile Detection and Excavation are left to the task specific section of the paper.

\subsection{State Estimation}
\label{subsec:localization}
Localization was one of the most challenging problems faced in this virtual environment provided by the competition, given that the robotic systems did not have a source of global localization, and due to many factors that led to drift in state estimates: high slippage, abundance of obstacles, low-featured lunar terrain, and variable lighting conditions. 

\begin{figure}[t!]
    \centering
    \includegraphics[width=0.95\linewidth]{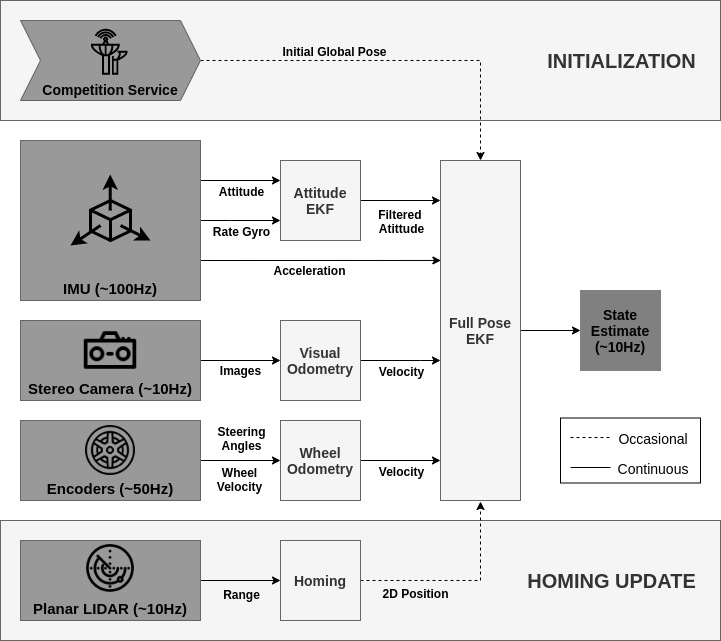}
    \caption{The architecture of the state estimation framework. Initialization was performed at the beginning and the Homing update was performed occasionally. The filtered attitude was obtained with an EKF dedicated to the provided IMU outputs. The velocity and the position of the rover were estimated in a sensor fusion EKF  using the wheel and visual odometry velocities. The full state of the rover is established by combining the filtered attitude from attitude EKF; velocity and position from the sensor fusion EKF.}
    \label{fig:localization2}
\end{figure}

The rover was equipped with wheel encoders, an IMU, a 2D LiDAR and a stereo camera, which were leveraged for localization. The simulated sensors had no bias but were disturbed by random noise. The noise was modeled as a zero-mean Gaussian distribution based on experiments. With the provided sensor package, the individual benefit of each sensor observation was leveraged using the extended Kalman filter (EKF) formulation:
\begin{equation}
\begin{aligned}
&\check{\mathbf{x}}_{k}=\mathbf{f}_{k-1}\left(\hat{\mathbf{x}}_{k-1}, \mathbf{u}_{k-1}, \mathbf{0}\right) \\
&\tilde{\mathbf{P}}_{k}=\mathbf{F}_{k-1} \hat{\mathbf{P}}_{k-1} \mathbf{F}_{k-1}^{T}+\mathbf{G}_{k-1} \mathbf{Q}_{k-1} \mathbf{G}_{k-1}^{T} \\
&\mathbf{K}_{k}=\check{\mathbf{P}}_{k} \mathbf{H}_{k}^{T}\left(\mathbf{H}_{k} \check{\mathbf{P}}_{k} \mathbf{H}_{k}^{T}+ \mathbf{R}_{k}\right)^{-1} \\
&\hat{\mathbf{x}}_{k}=\check{\mathbf{x}}_{k}+\mathbf{K}_{k}\left(\mathbf{y}_{k}-\mathbf{h}_{k}\left(\check{\mathbf{x}}_{k}, \mathbf{0}\right)\right) \\
&\hat{\mathbf{P}}_{k}=\left(\mathbf{I}-\mathbf{K}_{k} \mathbf{H}_{k}\right) \mathbf{P}_{k}.
\label{eq:EKF}
\end{aligned}
\end{equation}
where $\mathbf{x}_{k}$ is the state vector, $\mathbf{P}_{k}$, $\mathbf{F}_k$, $\mathbf{H}_k$, and $\mathbf{G}_{k}$ are the state uncertainty, state transition, measurement models, and noise Jacobian respectively. $\mathbf{Q}_k$ is the process noise covariance and $\mathbf{R}_k$ is the measurement noise covariance matrices. $\textbf{K}_k$ is the filter gain. The state estimation framework consisted of four-wheel steering (4WS) wheel odometry, visual odometry, attitude estimation EKF, periodic homing update~\cite{gu2018robot}, and sensor fusion EKF processes. The architecture of the implemented state estimation framework is depicted in Fig.~\ref{fig:localization2}. 

Instead of estimating the full state in a single EKF, we used two layers of filtering. The sensor fusion EKF estimated velocity and position but the attitude estimation was performed by another layer of filtering using the attitude EKF, which leverages both rate gyroscope measurements and noisy relative orientation measurements (i.e., changes in roll, pitch, and yaw relative to where the rover spawned). Given the specification of the simulated IMU provided by the competition, the estimated attitude was shown to be sufficiently accurate for the competition goals. Specific filter components for each state estimation model in (\ref{eq:EKF}) are provided for the sake of completion and reproducibility as follows. 
The state vector of the attitude EKF is 
\begin{equation}
\textbf{x}_{AF}=[\phi, \theta, \psi, p, q, r]^T
\end{equation}
where $\phi$ is roll, $\theta$ is pitch, $\psi$ is roll, and $p$, $q$, $r$ are the corresponding angular rates. The state transition model is established as:
\begin{equation}
\textbf{F}_{AF}=\begin{bmatrix}
F_{1,1} & F_{1,2} & 0  & -1  & F_{1,5}  & - \operatorname{sin}(\phi) \\ 
F_{2,1} & 1 & 0 &0  & - \operatorname{cos}(\phi) &\operatorname{sin}(\phi) \\ 
F_{3,1}  &F_{3,2}  &1 &F_{3,4}  &0  &F_{3,6} \\ 
 0&0 &0  &1  &0 &0 \\ 
 0&0  &0 &0 &1  &0 \\ 
0 &0 &0 &0  &0  &1 
\end{bmatrix} \Delta t_{IMU}
\end{equation}
where 
\begin{subequations}
\begin{alignat}{4}
  F_{1,1}=& \frac{1}{\Delta t_{IMU}} + \left(\operatorname{cos}(\phi) \operatorname{tan}(\theta)  q- \operatorname{sin}(\phi) \operatorname{tan}(\theta) r\right) \\
  F_{1,2}=& \frac{\operatorname{sin}(\phi)}{\operatorname{cos}^{2}(\theta)} q+\frac{\operatorname{cos}(\phi)} {\operatorname{cos}^{2}(\theta)}  r  \\
  F_{1,5}=& - \operatorname{sin}(\phi) \operatorname{tan}(\phi)\\
   F_{2,1} =&  \operatorname{sin}(\phi)  q + \operatorname{cos}(\phi) r \\
  F_{3,1} =& \frac{\operatorname{cos}(\phi)}{\operatorname{cos}(\theta)}  q - \frac{\operatorname{sin}(\phi)}{\operatorname{cos}(\theta)}  r \\
   F_{3,2} =& \frac{\operatorname{sin}(\phi)}{\operatorname{cos}^2(\theta)}  q + \operatorname{cos}(\phi) \operatorname{sin}(\theta)  r \\
  F_{3,4}=& \frac{-\operatorname{sin}(\phi)}{\operatorname{cos}(\theta)}\\
   F_{3,6} = &\frac{-\operatorname{cos}(\phi)}{\operatorname{cos}(\theta)}
\end{alignat}
\end{subequations}

The measurement model is given as 
\begin{equation}
\textbf{H}_{AF}= \begin{bmatrix}
\mathbf{I}_{(3\times3)},\mathbf{0}_{(3\times3)}
\end{bmatrix}
\end{equation}
and the covariance of the measurement noise is 
\begin{equation}
\textbf{R}_{AF}=\mathbf{I}_{(3\times3)}\sigma{_{m_{AF}}}^2
\end{equation}
where $\sigma{_{m_{AF}}}$ is the standard deviation of the measurement noise. 
% $\sigma{_{m_{AF}}}$ = 0.1
The covariance of the process noise is
\begin{equation}
\textbf{Q}_{AF}=\mathbf{I}_{(3\times3)}\sigma{_{p_{AF}}}^2
\end{equation}
where $\sigma{_{p_{AF}}}$ is the standard deviation of the process noise.
% where $\sigma{_{p_{AF}}}$ =0.05

After the initialization process (see Fig.~\ref{fig:localization2}), the state estimate in the local frame was transformed to the global frame by using a transformation matrix. The initial global position from the initialization is used as the initial state for sensor fusion EKF. The state vector for visual and wheel odometry velocity updates in the sensor fusion EKF is defined as
\begin{equation}
\label{eq:statevector}
\textbf{x}_{MF}=[R_x, R_y, R_z, Rv^N_x, Rv^N_y, Rv^N_z]^T
\end{equation}
where $R_{x}$, $R_{y}$, and $R_z$ are the rover location in global frame; $Rv_x$, $Rv_y$, $Rv_z$ are the rover's velocities for $x$, $y$, and $z$ axes in the navigation frame, respectively. The measurement innovation for visual odometry and wheel odometry velocity updates are
\begin{equation}
\textbf{z}_{O}=\textbf{C}_b^n[O_{Vx}, O_{Vy}, O_{Vz} ]^T, \quad  O\in(\text{WO}, \text{VO})
\end{equation}
where $C_b^n$ is the coordinate transformation matrix from the body frame to the navigation frame, $O$ is the measurement type (WO -- wheel odometry or VO -- visual odometry) that depends on the availability of that particular measurement. Note that camera data rate is 10\,Hz and encoder data rate is 50\,Hz, which makes the wheel odometry measurement updates 5 times faster than visual odometry measurement updates in the sensor fusion EKF. The measurement noise covariance matrix, process noise covariance matrix, and measurement model for these can be given as: 
\begin{equation}
\textbf{R}_{O}=\mathbf{I}_{(3\times3)}\sigma_{O}^2
\end{equation}
% $\sigma_{\text{WO}}$ = .05
%
\begin{equation}
\textbf{Q}_{O}=\begin{bmatrix}
  \sigma^2(\Delta t_{O})^4\mathbf{I}_{(3\times3)}& \mathbf{0}_{(3\times3)} \\ 
 \mathbf{0}_{(3\times3)} & \sigma^2(\Delta t_{O})^2 \mathbf{I}_{(3\times3)} 
\end{bmatrix}
\end{equation}
\begin{equation}
\textbf{H}_{O}= \begin{bmatrix}
\mathbf{0}_{(3\times3)},\mathbf{I}_{(3\times3)}
\end{bmatrix}
\end{equation}

% As mentioned before, sensor fusion EKF implementation did not estimate the attitude but output a full state estimate leveraging the external attitude estimation from the attitude EKF as is depicted in Fig.~\ref{fig:localization2}.

Notice that the fused estimation only leveraged the velocity estimations from the VO and WO, which, alone, were also able to estimate the rover pose. However, during testing, it is observed that the fused method has several advantages over VO and WO pose estimations. A localization accuracy comparison table of the localization methods is given in Table~\ref{tab:results1}. 

\begin{table} [htb]
\centering
\footnotesize
\begin{threeparttable}
\caption{Comparison of the Localization Methods}
\label{tab:results1}
\centering
\begin{tabular}{@{}lcccccc@{}}
\hline
   & \multicolumn{3}{c}{Abs. Error, x-axis (m)} & \multicolumn{3}{c}{Abs. Error, y-axis (m)}   \\
Run & \scriptsize{WIO\tnote{*}}& \scriptsize{VO\tnote{*}}& \scriptsize{VIWO\tnote{*}}& \scriptsize{WIO\tnote{*}}& \scriptsize{VO\tnote{*}}& \scriptsize{VIWO\tnote{*}}\\ 
\hline\hline
Seed\_19403	&81.74	&17.81	&\textbf{2.03}	&143.67	&\textbf{1.88}	&5.32\\
Seed\_19616	&7.44	&2.82	&\textbf{1.62}	&189.98	&2.72	&\textbf{0.78}\\
Seed\_25637	&77.03	&3.84	&\textbf{3.28}	&168.71	&\textbf{3.51}	&3.52\\
Seed\_27477	&72.94	&8.17	&\textbf{2.24}	&161.26	&4.28	&\textbf{2.73}\\
Seed\_33910	&13.73	&10.69	&\textbf{1.98}	&210.06	&5.62	&\textbf{0.86}\\
Seed\_39902	&\textbf{0.08}	&4.82	&1.56	&216.11	&1.76	&\textbf{0.21}\\
Seed\_98294	&166.85	&8.53	&\textbf{7.10}	&2.16	&2.09	&\textbf{0.17}\\
Seed\_1800	&167.61	&4.33	&\textbf{1.13}	&110.63	&\textbf{0.31}	&1.57\\
Seed\_1078	&144.22	&2.10	&\textbf{1.13}	&76.73	&17.64	&\textbf{2.11}\\
Seed\_1129	&123.39	&\textbf{4.32}	&11.73	&96.36	&\textbf{7.15}	&19.00\\
\hline\hline
STD	(m)        &64.14	&4.76	&3.42	&66.83	&4.97	&5.64\\
Average (m)	        &85.50	&6.74	&\textbf{3.38}	&137.57	&4.70	&\textbf{3.63}\\
Median	(m)        &79.39	&4.57	&\textbf{2.00}	&152.46	&3.11	&\textbf{1.84}\\

\hline
\end{tabular}
\begin{tablenotes}
\item[*] WIO: Wheel Odometry + IMU Heading, VO: Visual Odometry, \\VIWO: VO Velocity + IMU + WO Velocity.
\end{tablenotes}
\end{threeparttable}
\end{table}
\begin{figure}[t!]
    \centering
    \includegraphics[width=\linewidth]{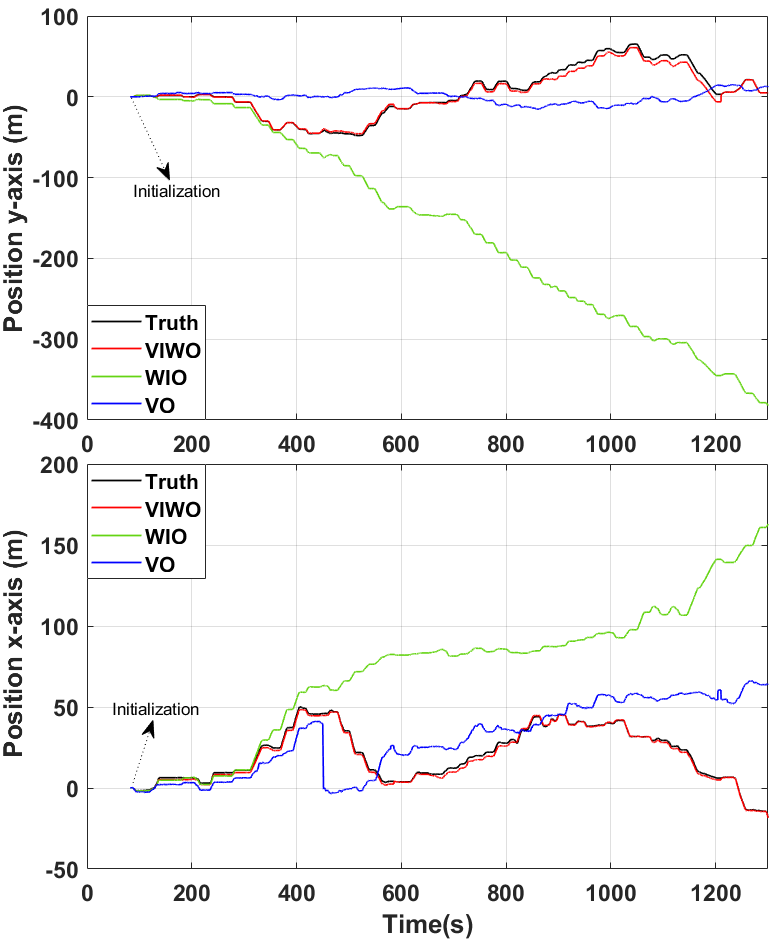}
    \caption{Horizontal localization accuracy comparison of tested localization methods in a typical run. Truth is shown as black line, wheel odometry with IMU heading aiding solution is the green line (WIO), visual odometry solution is depicted as a blue line (VO), and the visual, inertial, wheel odometry fused estimation is shown as red line (VIWO) for both axes. WIO solution inherently suffers from wheel slippage and it significantly drifts after a short drive. VO solution is accurate when there are sufficient features in the environment, but it suffers in feature-degraded areas. The fused estimation leverages both wheel odometry and visual odometry solutions along with IMU outputs in an EKF (Fig.~\ref{fig:localization2}) and provides a better estimate.}  
    \label{fig:LocEx}
\end{figure}
Due to wheel slippage, WO solution drifts significantly after a short drive even utilizing accurate IMU heading estimate. VO solution is accurate when there are sufficient features in the environment, but it generally fails while driving feature-degraded areas. Then, the fused estimation, which leverages both wheel odometry and visual odometry, provides a more reliable solution. A comparison of the localization accuracy given by visual odometry (VO), wheel odometry with IMU heading (WIO), and the proposed fused estimation (visual-inertial-wheel odometry -- VIWO) against the truth is shown in Fig.~\ref{fig:LocEx}.

As mentioned before, global localization estimation started with using a service provided by the competition, which reports the true pose of the rover with respect to the map. However, this could only be requested once per robot and per simulation. To get the maximum benefit from the knowledge of the robot's true pose, this capability was used to register the center of the processing plant as a global landmark, thus enabling future homing updates.  

Since the processing plant had a cylindrical shape, to register the base station as a global landmark, the 2D LiDAR data was used in a least squares estimator to fit the data to a circle, whose center was then registered in the global frame.
Note that registering the processing plant as a global landmark also assumed accurately known global attitude. 

Keeping the global attitude estimates reliable was one of the important aspects of our localization framework because it allowed the homing update to be performed more easily. In particular, by decoupling attitude estimation from position estimation when recognizing the processing plant for a homing update, because robot attitude could be assumed to be well-known, it was not necessary to estimate the orientation of the plant with respect to the robot, and only the processing plant's center location had to be estimated in the global frame.  The processing plant's center position estimates could then directly be used to correct position of the robots position solution by comparing them with their stored known global location. Given that $\check{PP}_{x}$, and $\check{PP}_{y}$ are the registered processing plant location in the global frame during the initialization, and using the state vector as in~(\ref{eq:statevector}), the measurement innovation for homing update can be expressed as 
\begin{equation}
\textbf{z}_{Homing}=\begin{bmatrix}
R_{x}+(\check{PP}_{ x} - \hat{PP}_{ x})\\ 
R_{y}+(\check{PP}_{ y} - \hat{PP}_{ y})\\ 
0\\ 
0\\ 
0\\ 
0
\end{bmatrix}
\end{equation}
where $\hat{PP}_{ x}$, $\hat{PP}_{ y}$ are the estimated processor plant's location. The effect of homing update to mitigate drifting errors will be reported and discussed in Section~\ref{sec:strategies}. 

Additionally, in order to keep the localization reliable, we adopted simple yet efficient innovation residual sanity checks to ensure that the measurements are consistent with the dynamics of the rover. These limits were heuristically determined with known state constraints of the rover.

In task 1, the mission constraints required at least 2\,\si{m} accuracy of the rover to provide resource location in the map. Since the volatiles could be detected by the dedicated volatile sensor at such a short range with respect to the rover, keeping a reliable and continuous localization solution for the rover played a critical role in correctly reporting the resource locations and reaching the desired waypoints for exploration.

In task 2, localization played an important role in reaching the resource location and being able to excavate it accurately. Since the pose service could be used once per rover, we were able to use the true pose service twice in this task, one for each rover. The Hauler's true pose service was used to mark the processing plant as a global landmark, similar to the resource localization task. To allow Excavator to reach the goal dig site before the Hauler, its true pose service was called when the Hauler updates its position, so the Excavator could start driving to the site. 
This occurred while the Hauler was performing the localization initialization phase with respect to the processing plant. In this task, the localization framework developed for task 1 was leveraged with an additional source of global localization updates. Specific to this task, the competition provided the global position of the resources, which means that any successful digging provides a way of correcting the localization drift. After arriving at the given resource location and successfully digging, the Excavator was able to update its pose estimate based on the given resource location. This was used in the localization framework as a pseudo-measurement update that leveraged the difference between the resource position estimate and the Excavator end-effector pose. Notice that it was assumed that we have precise yaw; thus, we decoupled roll/pitch estimations from yaw estimates, allowing the localization problem to be treated in the $xy$-plane. Successive failures of digging activities (i.e., not finding any resource in the area) indicate that the rover's localization is not reliable. In that case, the rover approached the processing plant for a homing update. 

In task 3, a global localization solution was only needed when the CubeSat position relative to the rover was estimated. In the period of looking for the CubeSat, a local dead-reckoning solution was used for the localization framework (i.e., spawning point assumed as the origin of the map). The rover initialized its global localization solution when the CubeSat position was estimated.

\subsection{Driving Control}
\label{subsec:driving}

Since the steering angles and torques of the wheels can be controlled individually, the rover was driven using a four-wheel steering (4WS) driving controller that uses different locomotion modes for pure translation, pure rotation, and combined translation and rotation. This approach was chosen because it reduces wheel slip with the terrain when compared to the skid steering driving controller provided by the competition. In our approach, the locomotion mode was selected based on the desired forward speed and rotational speed of the robot.

Depending on the values of these speeds, the 4WS driving controller decided which locomotion mode to use. When both components of the input command were required, e.g., during the traverse, the robot moved using a double Ackermann locomotion mode. For pure translational motion, synchronous-drive mode (crab motion) was used, and for pure rotation on the z-axis, point turns (turn-in-place maneuver) were used. All these locomotion modes can be found in further detail in~\cite{gounaris2011lunar}. As a result, the 4WS driving controller output individual steering angle and wheel velocity commands for each wheel. The desired wheel velocities were controlled using a simple proportional~(P) controller. 

The competition also provided a braking service, with the option of braking from 0 to 100\% where 100\% would lead to a braking limit of 500\,\si{Nm/rad} to each wheel simultaneously. As an alternative to the provided braking service, another braking option was included by setting the wheel speeds to zero to prevent the rover from slipping while trying to stop on slopes.

\subsection{Navigation}
\label{subsec:nav}

Even for the simple task of traversing from point A to point B, many decisions need to be made by an autonomous rover to ensure its safe and efficient completion.
Navigation was done using the Move Base~\cite{movebase} framework, which uses a global planner to generate a global path between two waypoints and, a local planner to generate velocity outputs to follow the global path as closely as possible, considering the vehicle dynamics. The local planner also takes into account a local costmap, created in real-time from obstacle point clouds and represented as a 2D occupancy grid. The Move Base framework provides a broad range of global and local planner implementations that can be selected based on specific mission requirements. Finally, the package also permits configuring recovery behaviors such as turn-in-place and clearing costmaps if the planners fail to find plans due to unexpected events.

Our choice of global and local planners were based on the literature evaluation of computational needs and plan execution performance metrics. Additionally, easiness of installation, documentation availability, and usage flexibility metrics were used to assist our decision. These criteria are summarized and compared in Table~\ref{table:planners}. For the global planner, Base Global Planner was chosen because it is faster than the NavFn. It is also and more flexible and reliable than the Carrot Planner. All global planners were easy to install and had sufficient documentation. The chosen local planner was dynamic window approach (DWA) Local Planner because it considers the dynamics of the robots, and provides the highest computational efficiency and similar execution performance to the other methods~\cite{fox1997dynamic,rosmann2017integrated,gerkey2008planning}.

\begin{table}[htb!]
\centering
\caption{Comparison of Move Base Planners }
\begin{threeparttable}
\begin{tabular}{@{}lcccccc@{}} 

    \hline
    \hline
\textit{Global Planners} & CE & EP  & EI & DA & UF & Reference \\ 
   \hline
navfn   & *  & ** & **  & *  & * & \cite{brock1999high}\\
\textbf{base } & ** & ***& *** & ** & ** & \cite{brock1999high}\\
carrot	& ***& *  &	***	&*	 &*  & \cite{movebase}\\
    \hline
    \textit{Local Planners} & CE & EP  & EI & DA & UF & Reference\\ 
       \hline
base    & ***	&**	&***	&*	&* & \cite{gerkey2008planning, fox1997dynamic} \\
\textbf{dwa	}    & ****	&**	&***	&***	&*** & \cite{gerkey2008planning, fox1997dynamic,filotheou2019quantitative}\\
eband   & **	&***	&**	&*	&**& \cite{quinlan1993elastic}\\
teb	    & **	&*** &*** &	***&	*** &\cite{rosmann2017integrated,filotheou2019quantitative}\\
mpc	    & *	&**	&**	&*	&** & \cite{rosmann2020online}\\
 \hline
 \hline
\end{tabular}
\begin{tablenotes}
\item CE: Computational Efficiency, EP: Execution Performance, EI: Easiness of Installation, DA: Documentation Availability, UF: Usage Flexibility.
\end{tablenotes}
\end{threeparttable}
\label{table:planners}
\end{table}

Our navigation strategy is based on creating a sequence of waypoints off-line and using Move Base to move from a waypoint to another. Point clouds relative to obstacles (see Section~\ref{subsec:cv}) were used on-line to check if waypoints were achievable by comparing them with the coordinates of the desired waypoint. If a waypoint existed inside the region defined by the point cloud, it was rejected, and the next waypoint was requested. If it existed outside the cloud, the robot proceeded to a planning stage which generated a navigation plan for that goal.

For task 1, approximately one hundred waypoints were used for each round because it was observed that the rover could not reach more than that in each 45 minute round even when driving at the maximum speed. To generate the waypoints, the overall map was segmented into five regions, representing each of the map's crater areas. The exploration of these five regions was then prioritized based on their traversability  (e.g., steepness of crater slope and surrounding terrain) and the likelihood of volatiles being present. To assess the likelihood of volatile locations, a Gaussian mixture model (GMM)~\cite{reynolds2009gaussian}, that was estimated using volatile locations from prior simulation trials, was developed. The GMM was then sampled to form a candidate group of waypoints that were connected to form exploration routes covering each of the five map regions. Each exploration route was designed to both start and end close to the center of the map.  By ending an exploration route at the center of the map, the robot would be in proximity to the processing plant such that a homing update could be performed before the robot enters the next exploration plan.  Figure~\ref{fig:localization} shows examples of three exploration routes.

For task 2, volatile locations were provided by the competition. Instead of having a waypoint generator, this time, one of the volatile locations was selected based on the flatness of the area surrounding it and the distance from the current Excavator position. Then, this goal was sent to the navigation stack, which provides a path for both the Excavator and Hauler. Once the Excavator and Hauler reached the proximities of the goal, the Excavator stopped at 1\,\si{m}, and the Hauler stopped at 4\,\si{m} from the target volatile. Then, the two robots proceeded to the excavation phase as detailed in Section~\ref{subsec:round2}.

For task 3, a more straightforward random search strategy was adopted without using the Move Base framework. For this search strategy, the robot moved in random directions out from the base and returned to the processing plant repeatedly, ensuring that the robot explored the environment evenly and safely. This strategy maximized the chances to find the CubeSat which was randomly placed around the base station. The risks of getting lost or stuck in a crater increase with longer drives to the edges of the map; therefore, the regions closer to the center of the map were prioritized. However, the robot took more risks and drove long distances to try to find the CubeSat closer to the time limit of the simulation. 

\subsection{Object Detection}
\label{subsec:cv}
% The vision requirement is that obstacles and objects that are important to the competition should be identified in the image and their position known in the world.

A vision module was used to identify important objects and to localize their position with respect to the environment. The images provided by the stereo camera pair had 640x480 resolution and 10\,fps with artificial noise added to them. To detect, classify, and estimate the position of these features, a deep learning based algorithm that solves the three steps in a single pass was the ideal solution. Among state of the art object detection methods, the Single Shot MultiBox Detector (SSD)~\cite{liu2016ssd} and YOLO~\cite{redmon2017yolo9000} are methods that are able to perform inference at more than 10\,Hz, which allows to process all the image data obtained from simulation in real time. Both of these methods present similar accuracy, measured as the intersection over union (IoU) values of a network trained on the same datasets. Currently, there are a few versions of each of these algorithms. The VGG16 based SSD network~\cite{liu2016ssd} was chosen because it is trainable with a smaller dataset of labelled images. The single-shot multi-box detector~(SSD) based on the VGG16 architecture~\cite{simonyan2014very} returns a bounding box around each detected object, its probable class, and the confidence of that inference. In our implementation, transfer learning was used from a network previously trained in the Microsoft common objects in context (MS COCO) dataset~\cite{lin2014microsoft}, which consisted of $350,000$ images and $80$ object categories.

\begin{figure}[t!]
    \centering
    \includegraphics[width=\linewidth]{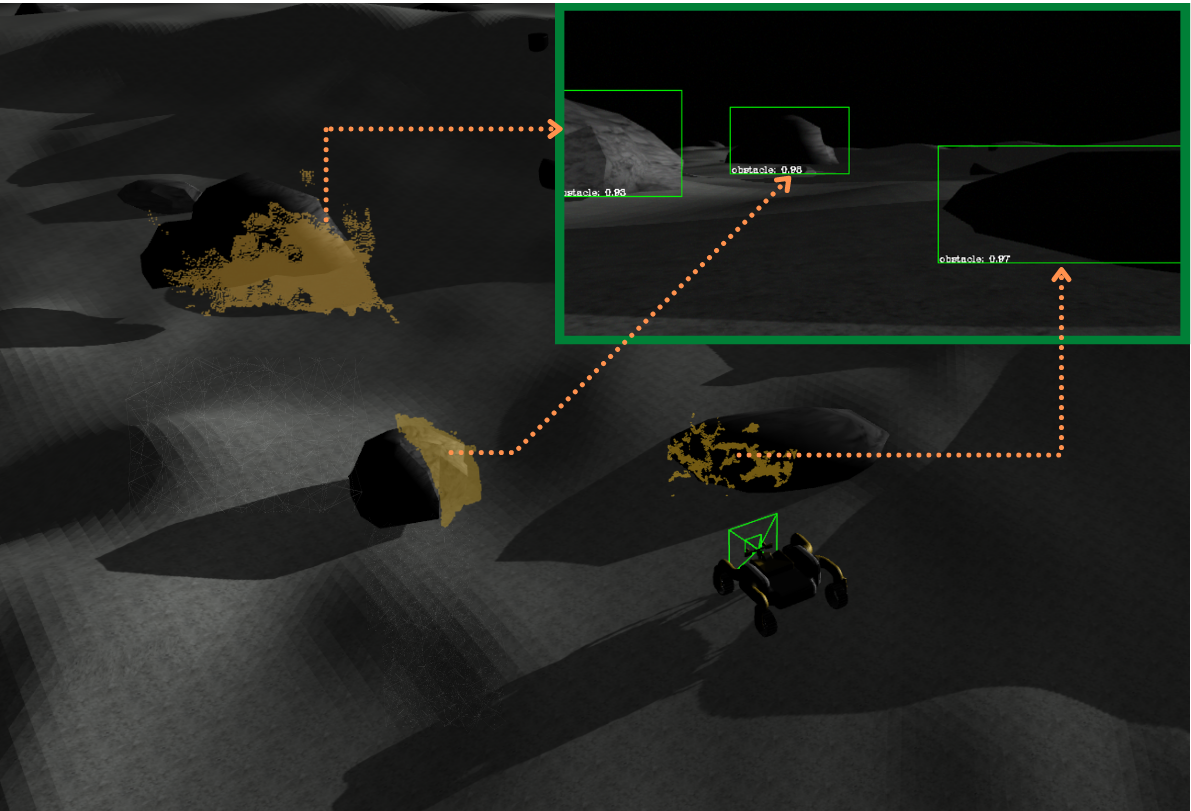}
    \caption{Obstacle detection and point cloud segmentation. On the top right, the obstacles in front of the robot were detected in real time using a single shot multi-box detector network, resulting in the green bounding boxes. These bounding boxes and the disparity images came from the rover stereo camera were also processed to obtain the segmented point clouds of the obstacles, which are shown in yellow.}
    \label{fig:obstacles}
\end{figure}

The SSD network was designed to identify seven classes:  processing plant, CubeSat, fiducial marker on the processing plant, rocks in the environment,  visible volatiles above or partially above the surface, craters, and the other rovers. The new architecture had 24,547,000 trainable parameters. Images obtained from driving the robot in the simulation environments were used. These images were randomly sampled from more than four hours of robot operation. They were manually labeled with 1500 images used as training data and 500 images for validation. 

The bounding box for each detected object was used to extract its 3D point cloud information. First, the disparity image from the stereo camera pair was calculated using semi-global block matching~\cite{hirschmuller2005accurate}. Then, using the bounding box coordinates, each pixel inside the object bounding box was used to calculate the 3D point cloud. 

Each point was considered valid only if there was a feature match in the disparity image pixel value. Due to the limited size of the environment and the resolution of the camera, $z$ coordinate (optical axis) values larger than a given threshold (1000~\si{m} in our case) were also discarded. If the calculated $z$ was valid, the bounding box with disparity information was used to estimate the full 3D coordinates of each point relative to the rover camera frame. Figure~\ref{fig:obstacles} shows examples of rocks seen by the camera in front of the robot and the processed point cloud with semantic information about these obstacles. 

For task 1, SSD was used to identify the obstacles as point clouds that were used for general obstacle avoidance along its path. For that, the point cloud was clustered to differentiate between different rocks detected. Centroids of each of the clusters were determined to get an idea of the height of the obstacles. Planes were also fitted to each of the clusters and normal vectors were calculated. The normal vectors represent the steepness of the obstacles. Since there were some point clouds detected as obstacles (e.g., crater slopes and small rocks), which were traversable, thresholds were set for both the centroid height (0.1\,\si{m}), and angle of the normal from the vertical (5\,\si{deg}). The clusters with properties higher than these two thresholds were put together to form a filtered version of the previous point cloud and published as obstacles. They were also converted to costmaps for Move Base. The images were also used to find and approach the processing plant for localization filter homing updates.

For task 2, an additional capability for the Excavator and the Hauler to detect each other was included. The position of one robot can be estimated from the other by averaging the point cloud values. This capability also enabled the Hauler to use visual servoing to find and approach the Excavator.

\subsection{Manipulation}

The Excavator and Hauler rovers were assigned with the mission of collecting resources from the terrain. For that, the Excavator had to drive to known locations and use its robotic arm to dig volatile substance and drop it into the Hauler's bin. In order to reduce computational complexity and to perform digging actions within the mission time limit, we opted for a simple approach where the kinematic equations were used to help controlling the motion of the arm from predefined configurations to the target points (volatile position or Hauler's bin). 

\begin{figure}[t!]
    \centering
    \includegraphics[width=\linewidth]{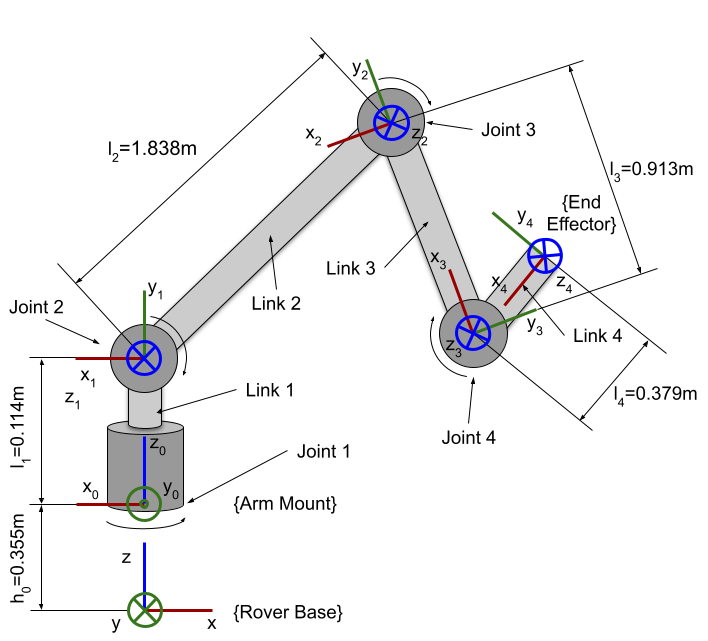}
    \caption{Excavator's arm coordinate frames, links and joints. Joint 1 corresponds to shoulder yaw, joint 2 to shoulder pitch, joint 3 to elbow pitch, and joint 4 to wrist pitch. Link 4 represents the bucket that is used to scoop the volatiles from the terrain.}
    \label{fig:SRC2_schematics}
\end{figure}
\begin{figure}[t!]
    \centering
    \includegraphics[width=0.9\linewidth]{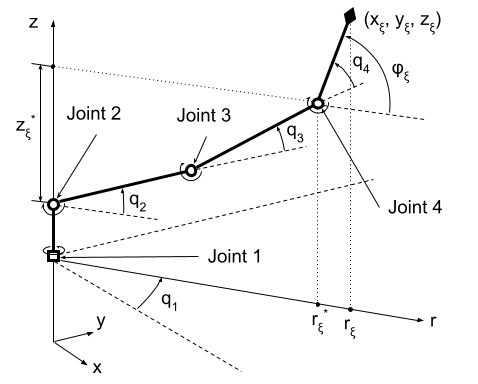}
    \caption{Geometric representation of the Excavator's arm. The end-effector pose is given by a position $P_\xi = [x_\xi, y_\xi, z_\xi]$, and a pitch angle, $\varphi_\xi$ with respect to the mobile base coordinate frame (see Figure~\ref{fig:SRC2_schematics}).}
    \label{fig:src2_ik}
\end{figure}
First, coordinate frames were assigned to each joint where demonstrated in Fig.~\ref{fig:SRC2_schematics}. Then, Denavit-Hartenberg parameters were obtained as shown in Table \ref{tab:src2_dh}. With this information, forward and inverse kinematics relations were derived for the arm.

\begin{table}[htb!]
\centering
\caption{Denavit-Hartenberg Parameters for \\ NASA's SRC2 Excavator Manipulator}
\begin{tabular}{@{}ccccc@{}}
\hline
Joint i & $a_i$ [m] & $\alpha_i$ [rad] & $d_i$ [m] & $\theta_i$ [rad] \\ 
\hline
\hline
1 & $0.0$ & $0.0$ & $l_1$ & $q_1$ \\
2 & $-l_2$ & $\pi/2$ & $0.0$ & $q_2$ \\
3 & $-l_3$ & $0.0$ & $0.0$ & $q_3$ \\
4 & $-l_4$ & $0.0$ & $0.0$ & $q_4$ \\
\hline
\end{tabular}
\label{tab:src2_dh}
\end{table}

The forward kinematics formulation is straight-forward for this 4R manipulator and the equations are obtained geometrically from the link lengths and joint angles defined in Figs.~\ref{fig:SRC2_schematics} and~\ref{fig:src2_ik}. The equations for the relative position and the bucket angle are given by:
\begin{subequations}
\begin{alignat}{4}
    \varphi_\xi &= q_{2} + q_{3} + q_{4} \\
    x_\xi &= r_\xi \cos(q_{1}) \\
    y_\xi &= r_\xi \sin(q_{1})\\
    z_\xi &= h_0 + l_1 + l_2 \sin(q_{2}) + l_3 \sin(\varphi_\xi-q_{4}) + l_4 \sin(\varphi_\xi)
\end{alignat}
\end{subequations}
where $r_\xi = l_2 \cos(q_{2}) + l_3 \cos(\varphi_\xi-q_{4}) + l_4 \cos(\varphi_\xi)$ and the $\xi$ subscript refers to a reference point at the end-effector (bucket). With this equations, given some input joint angles $q_{input} =  [q_{1}, q_{2}, q_{3}, q_{4}]$, it is possible to obtain the pose of the end-effector with respect to the mobile base, as shown in Fig.~\ref{fig:src2_ik}.

The inverse kinematics was also obtained geometrically by using two orthogonal, uncoupled planes of motion: one considers changing the azimuth of the whole arm (shoulder yaw), and the other considers changing the configuration of the arm (shoulder pitch, elbow pitch, wrist pitch) in the $z$-$r$ plane shown in Fig.~\ref{fig:src2_ik}. The required joint angle $q_1$ is obtained directly using the cylindrical coordinates and the other joint angles are obtained using the method described in~\cite{kumar2018} for a 3R planar manipulator. The equations for the four joint angles are:
\begin{subequations}
\begin{alignat}{4}
    q_1 &= \text{atan2}\left(y_\xi, x_\xi \right) \\
    \begin{split}
    q_2 &=  \text{atan2}\left(\frac{-z^{\ast}_{\xi}}{\sqrt{r^{\ast 2}_{\xi} + z^{\ast 2}_{\xi}}}, \frac{-r^{\ast}_{\xi}}{\sqrt{r^{\ast 2}_{\xi} + z^{\ast 2}_{\xi}}} \right)\\ 
    &\pm  \text{acos}\left(\frac{-(r^{\ast 2}_{\xi} + z^{\ast 2}_{\xi} + l_2^2 - l_3^2)}{2 l_2\sqrt{r^{\ast 2}_{\xi} + z^{\ast 2}_{\xi}}}\right)\end{split}\\
    q_3 &= \text{atan2}\left(\frac{z^{\ast}_{\xi}-l_2\sin(q_2)}{l_3},\frac{r^{\ast}_{\xi}-l_2\cos(q_2)}{l_3}\right) - q_2\\
    q_4 &= \varphi_\xi - (q_2 + q_3)
    \end{alignat}
\end{subequations}
where $ r_\xi = \sqrt{x_\xi^2 + y_\xi^2}$, $r^{\ast}_{\xi} = r_\xi - l_4 \cos(\varphi_\xi)$, and $z^{\ast}_{\xi} = z_\xi - h_0 - l_1 - l_4 \sin(\varphi_\xi)$. With these equations it is possible to obtain the required joint angles $q_{required} = [q_1, q_2, q_3, q_4]$, given a desired position $P_{desired}$ and angle $\varphi_{desired}$ for the bucket.

After solving the Excavator's arm forward and inverse kinematics there are several ways to plan its motion. The planning constraints included: 1) avoiding collisions and 2) maintaining the bucket's global angle within a specific range to ensure that the volatile was collected from the terrain and not dropped unintentionally. Predefined configurations were selected to act as intermediate waypoints for the arm to guarantee that there will be no collision during the motion. Thus, trajectories were obtained by interpolating joint angles in between the waypoints. More details of the excavation procedure are given in Section~\ref{subsec:round2}, where we discuss task~2's strategy.

\section{Task Strategies}
\label{sec:strategies}
\begin{figure}[t!]
    \centering
    \includegraphics[scale=0.35]{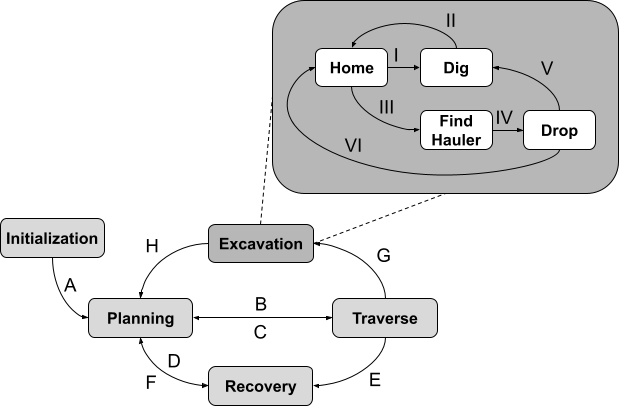}
    \caption{State machine architectures for the robots in tasks 1 and 2. The events that trigger transitions between states are given by the capital letters from \textbf{A} to \textbf{H}. For the system-level state machine, \textbf{A)} was triggered after obtaining the true pose from the global localization service; \textbf{B)}, when block planning generated a traversable navigation plan; \textbf{C)}, upon arrival at a desired waypoint location; \textbf{D)}, when a navigation plan could not be generated. \textbf{E)}, when the rover experienced motion issues; \textbf{F}), when the rover successfully performed recovery procedures and was able to reset its navigation plan; \textbf{G)}, when Excavator and Hauler arrived at an estimated location close to a target volatile, which enabled its secondary state machine dedicated to excavation; and \textbf{H)}, when the full mass of the volatile was collected, or if the excavation was not successful in a predetermined time window. The sequence of actions of the excavation state machine is ordered from I to VI.}
    \label{fig:nested_smach}
\end{figure}

As discussed in the Section~\ref{sec:overview}, the qualification round consisted of three tasks. Each task had to be completed within a simulation time limit of 45 minutes.

The virtual lunar environment contained hills, slopes, rocks, craters, including some that lead to permanently shadowed regions, volatile resources, and a processing plant. There were eight different volatile types in the environment, which were randomly distributed, and mostly under the surface of the lunar environment. Large lunar rocks acted as obstacles and were distributed in random locations for each seed of the simulation. The processing plant was a simulated lander that also spawned with a randomized pose, close to the center of the environment. This environment is shown in Fig.~\ref{fig:environment}.

To execute the task-specific strategies, a centralized autonomy approach for decision making was used in the first two tasks. For that, a state machine framework that autonomously decides the actions of the rover during the exploration was developed. 

\subsection{Task 1 -- Resource Localization}
\label{subsec:round1}

\begin{figure}[t]
    \centering
    \includegraphics[width=\linewidth]{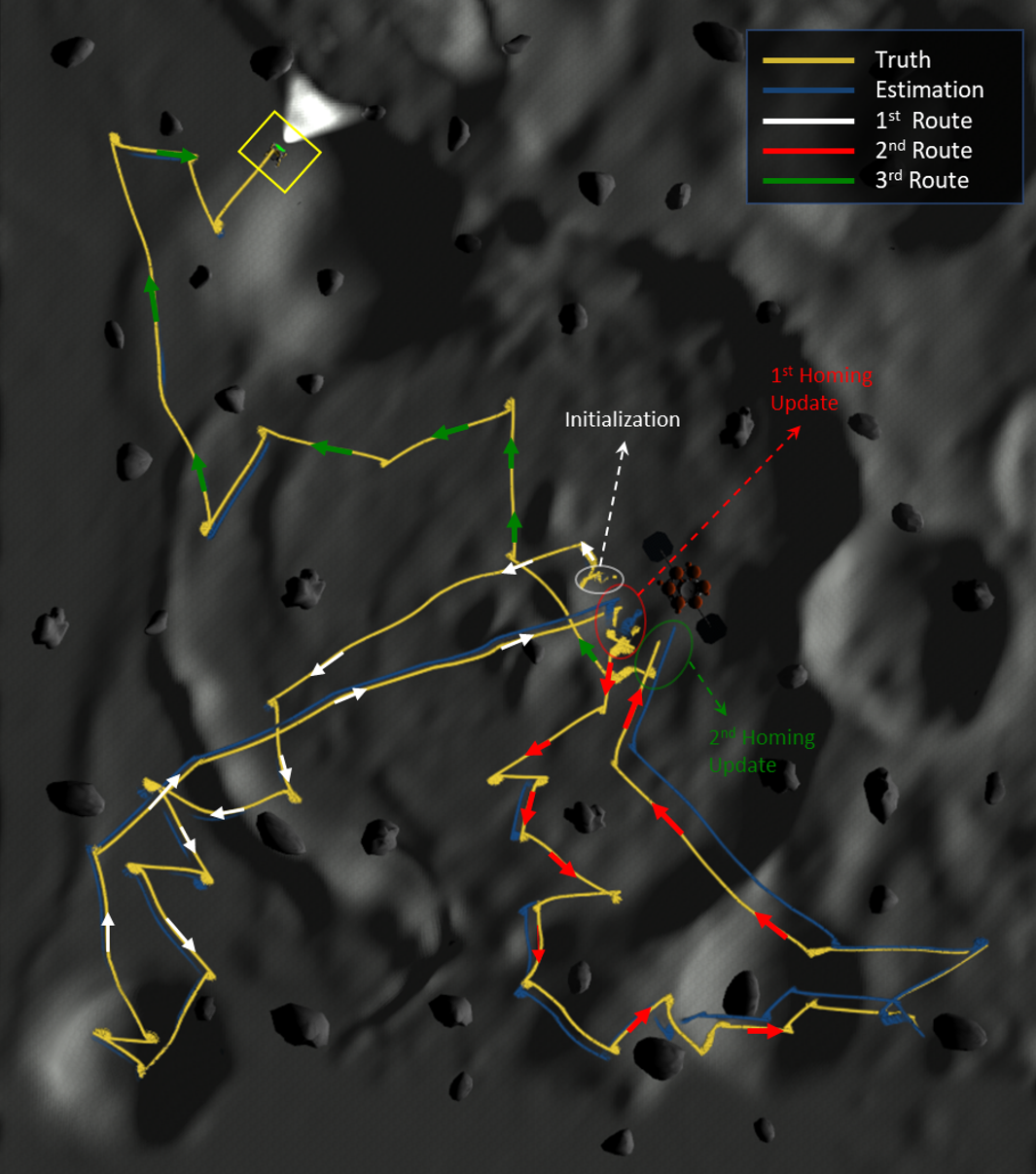}
    \caption{An example of a simulation run for task 1. The rover, first, is initialized with the service provided by the competition. After that, the rover drives to the previously generated waypoints to explore volatiles. This figure presents three routes. White arrows show the first route, which starts after the initialization, red arrows show the second route, which starts after the first homing update, and green arrows show the third route, which starts after the second homing update. The yellow line indicates the truth odometry, and the blue line indicates the estimated localization solution. After each homing update, the rover reset the localization solution with respect to the base station. As shown in the figure, after each update, the truth and estimation lines became almost aligned. These homing updates were significantly useful to keep localization estimation accurate within 2\,\si{m} radius.}
    \label{fig:localization}
\end{figure}

The state machine used to control the robot in task 1 is shown in Fig.~\ref{fig:nested_smach}. In the initialization state, the robot detected and approached the processing plant as detailed in Sections~\ref{subsec:cv} and~\ref{subsec:nav}, respectively. Using the provided true pose service, the robot estimated the processing plant location and recorded it as a global landmark as described in Section~\ref{subsec:localization}. The rover state transitioned to planning upon completion of the initialization phase. In the planning state, the robot requested a waypoint, which was provided by a waypoint generator node as described in Section~\ref{subsec:nav}. If there was no collision in its field of view, the waypoint was passed to the navigation framework, and the state was transitioned to traverse. In the traverse state, the rover drives from one waypoint to another by using navigation and driving subsystems as detailed in Sections~\ref{subsec:nav} and~\ref{subsec:driving}, respectively. An example of the localization solution that demonstrates the waypoints and routes along with the drifts during the exploration task is given in Fig.~\ref{fig:localization}. 

\begin{table*} [h!]
\centering
\footnotesize
\begin{threeparttable}
\caption{Comparison of volatile reporting accuracy and mitigated error values with homing strategy for 10 random simulation seeds.}
\label{tab:results}
\centering
\begin{tabular}{@{}lccccccccc@{}}
\hline
{} & \multicolumn{1}{c}{} & \multicolumn{6}{c}{Corrected Error, Horizontal (m)}  & \multicolumn{2}{c}{Volatile Report }\\
Run & \scriptsize{\# of Homing}& \scriptsize{Homing \#1}& \scriptsize{Homing \#2}&\scriptsize{Homing \#3}&\scriptsize{Homing \#4}&\scriptsize{Homing \#5}&\scriptsize{Homing \#6}&\scriptsize{Sensed}& \scriptsize{Scored}\\ 
\hline\hline
Seed \#1800	&4	&0.32	&13.23	&1.28	&0.57	&N/A	&N/A	&14	&12\\
Seed \#1078	&4	&1.95	&8.04	&0.30	&4.90	&N/A	&N/A	&10	&10\\
Seed \#19403	&4	&3.77	&19.61	&2.22	&13.89	&N/A	&N/A	&11	&9\\
Seed \#1129	&4	&22.20	&3.14	&5.33	&7.49	&N/A	&N/A	&6	&5\\
Seed \#98294	&4	&6.93	&3.79	&2.84	&2.98	&N/A	&N/A	&6	&5\\
Seed \#39902	&4	&1.46	&3.99	&11.26	&7.70	&N/A	&N/A	&5	&4\\
Seed \#27477	&5	&3.37	&2.33	&0.83	&4.89	&10.11	&N/A	&9	&9\\
Seed \#33910	&5	&0.92	&-0.94	&6.99	&11.71	&7.01	&N/A	&9	&9\\
Seed \#19616	&5	&1.60	&3.00	&5.93	&12.68	&5.69	&N/A	&5	&4\\
Seed \#25637	&6	&4.23	&4.93	&2.67	&4.39	&13.65	&7.83	&8	&8\\
\hline
\end{tabular}
\end{threeparttable}
\end{table*}
\begin{figure*}[htb!]
    \centering
    \includegraphics[width=\linewidth]{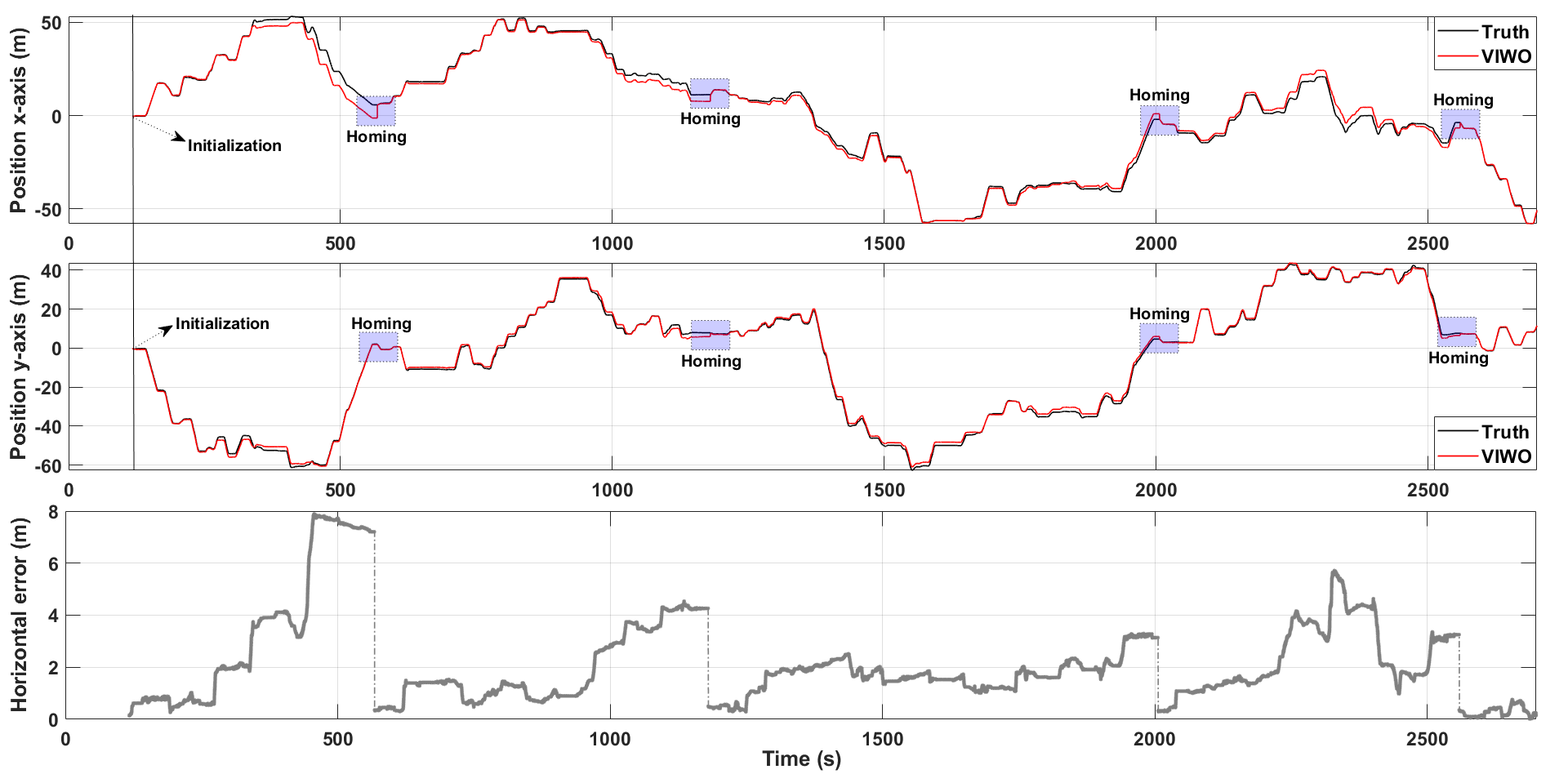}
    \caption{Horizontal localization accuracy of the fused estimation (VIWO) and impact of using homing strategy for Seed \#98294. This strategy is used effectively as a loop-closure technique and  significantly reduces the localization error when used.} 
    \label{fig:LocSeed}
\end{figure*}
The recovery state aimed to minimize the possible failures in planning and traverse states.
If the rover was experiencing immobility issues, the recovery state was triggered by the following indicators:  1) excessive slip detection; 2) steep slope detection; 3) stuck detection. Excessive slip detection was inspired by the approach in~\cite{improvedSlip}, which detected high slippage using discrepancies between VO and wheel odometry velocities. Steep slope detection used a heuristically determined threshold for climbing limits (e.g., 35 deg) of the rover by using the rover pitch angle estimates to minimize the rollover. Stuck detection used 2D LiDAR to determine if the rover was not able to move due to an obstacle in front of it. After triggering any of these indicators during the traversal, the rover executed predetermined maneuvers to regain its mobility. Also, when the navigation plan was not achievable in the planning state, the rover reset its current plan and changed its state to the planning state to generate a new navigational plan. 

When a volatile was sensed during driving, the rover reported the location of the volatile using its own localization solution while considering the lever arm of the mounting location of the volatile sensor with respect to the IMU. For volatile reporting, the rover used the logic to anticipate the volatile position as detailed in  later in this section. After visiting a number of predetermined locations, the rover proceeded to a homing phase. In this phase, the rover drove to the processing plant, approaching it with visual servoing, and then performed a localization update.

The error mitigation after performing the localization update (homing) is shown in Table~\ref{tab:results}. Note that without having a loop-closure strategy like this homing update, the localization error would significantly increase due to wheel slippage and VO failures in this steep-sloped, low-featured environment. Consequently, any localization inaccuracy issue in the early stages of the simulation run would yield consecutive unsuccessful reports for the sensed volatiles. Even some of the homing updates would seem redundant in a manner of localization error mitigation, these updates provided a considerable assist to keep the rover's localization accuracy sufficient to score the sensed volatiles with a success rate of more than 80\%.

Additionally, data from one of the simulation runs with positioning estimate against truth and horizontal error mitigation is illustrated in Fig.~\ref{fig:LocSeed}. The rapid increase in the horizontal error is most likely due to struggling against a high-slip environment (e.g., climbing up/down a steep-slope crater) or an unexpected rover stuck due to obstacle avoidance failure. However, the rover was able to recover its localization accuracy (sub-meter level) after a homing update. 

\paragraph*{\textbf{Volatile Handling}}

In parallel to all executed tasks in the state machine, after the volatile sensor was triggered for the first time, the rover started computing volatile location estimations. Every time the sensor was triggered again, a new volatile with its ID, type, and the current rover location estimation was queued. In our design, the rover did not stop when a volatile is sensed. Instead, we recorded the estimated volatile position and let the rover anticipate the location of the volatile while driving. Since the volatile sensor continuously reported a volatile until it was scored, our volatile handling strategy had to overwrite the volatile sensor report such that only the location in which the sensor was closest to the volatile was stored for attempting a scoring report. This was done to give the highest chance for scoring since both the sensor's detection range and report accuracy threshold were 2\,\si{m}. To enable this overwrite, the possibility of reporting a volatile was disabled while a volatile was actively being sensed. Once a volatile was no longer being actively sensed, its closest location was stored. The volatile score reporting service provided by the competition was given a randomized minimum delay timeout of 15-30\,\si{s}. Therefore, a reporting strategy was developed that first sends the current best estimate of the rover localization solution after accounting for the volatile sensor's lever arm. If the initial reporting based on the current best estimate was successful, the volatile was not queued. Otherwise, the volatile was queued for further scoring attempts. 

To account for the limited number of scoring attempts that could occur due to the imposed timeout, scoring attempts of a queued volatile were given lower priority with respect to newly sensed first attempt volatiles.  Additional attempts for queued volatiles were estimated using two different approaches.  For the first approach, each time a homing update was performed to correct the localization solution of the rover, all of the queued volatiles that were accumulated since the previous homing update had their locations corrected using the estimated localization drift. Second, in the case that no other volatiles were available for scoring, a simple search pattern around the vicinity of the estimated rover position was attempted that considered the rover's global heading at the time of sensing and the 2\,\si{m} accuracy threshold. By separating exploration from volatile handling, the robot was able to cover as much area as possible on the map in the 45\,minutes mission.

\subsection{Task 2 - Resource Collection}
\label{subsec:round2}

Our approach for this task was a direct extension of the approach used for the resource localization task and also rely on the state-machine of Fig.~\ref{fig:nested_smach}. An initialization step set the global reference frame and established the processing plant as a landmark on the map for future homing and localization recovery. Next, the robots would decide the next goal, plan a route, and move towards the selected volatile location. The Excavator parked directly in front of the volatile and waited to start excavating, then the Hauler parked at a short distance behind the Excavator. Once the excavation started, the Excavator executed a set of maneuvers to find the volatile, and the Hauler received a command to perform visual servoing to approach the Excavator based upon the computer vision detector. Whenever a volatile was found, its position, which was known \textit{a priori}, was used to update the Excavator's localization estimate. After collecting the entire mass of the volatile, the robots transitioned to the planning phase, selected a new goal, and repeated the process.

\paragraph*{\textbf{Excavation}}

The excavation phase was responsible for computing a trajectory for the manipulator. The Excavator's arm needed to dig the volatile from the terrain and drop it in the Hauler's bin. This needed to be performed at least twice for each resource in the map, because the Excavator's bucket could only carry up to half of the total resource mass per scoop.

\begin{figure}[t]
    \centering
    \includegraphics[width=\linewidth]{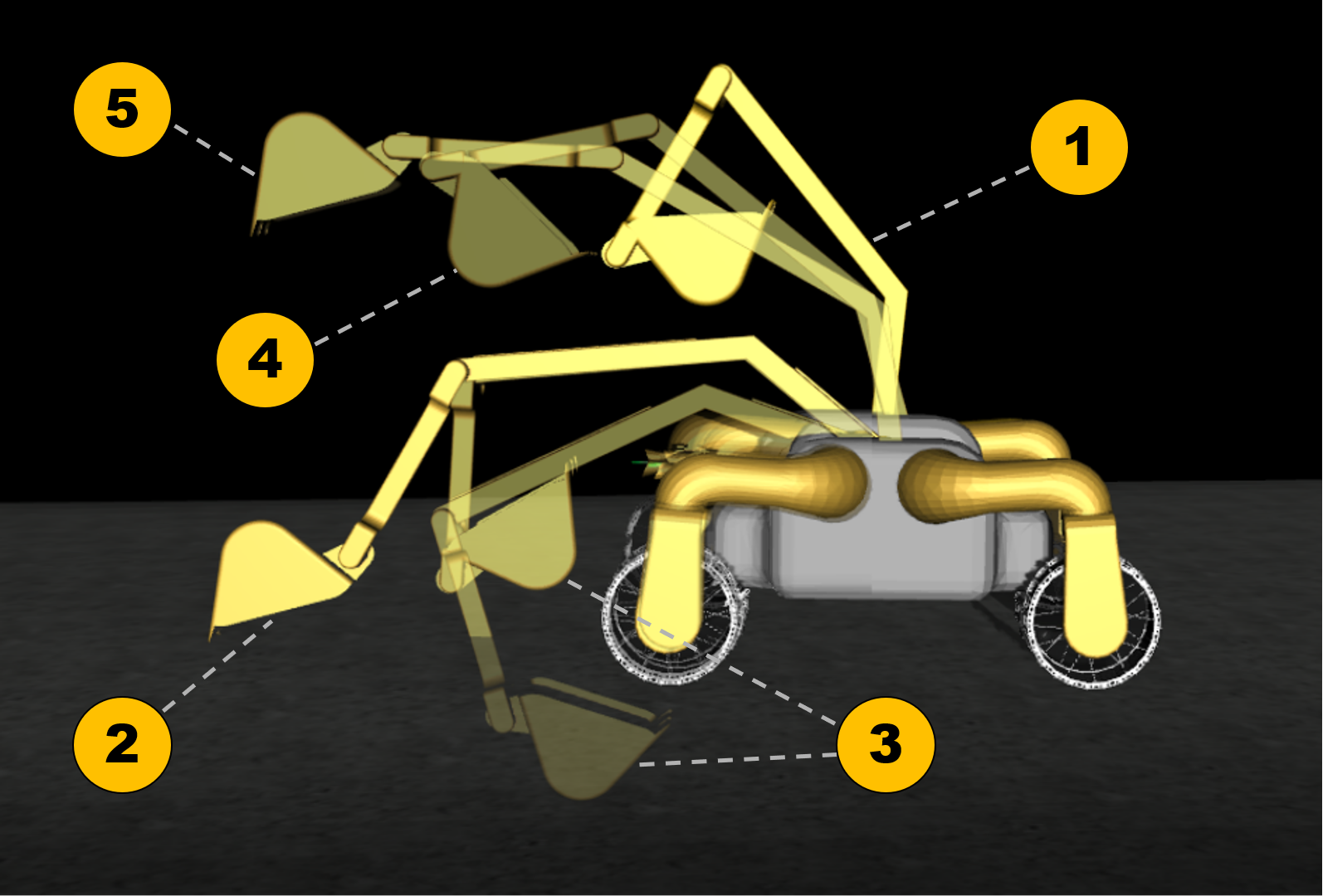}
    \caption{Predefined configurations of the Excavator. Home position for the arm is denoted as 1. This was used as an intermediate configuration between digging and dropping and also as the configuration for cruising. Configurations 2 and 3 were related to digging. In 2, the robotic arm was lowered to the ground level, and in 3, the arm scooped the terrain to excavate the volatile. Configurations 4 and 5 were related to dropping the volatiles in the Hauler's bin. In 4, the arm was extended, and in 5, the bucket was rotated, dropping its contents. All the configurations can be performed with adjustable headings.}
    \label{fig:ManipulationStates}
\end{figure}

Once the Excavator reached the location close to the volatile that needed to be excavated, it enabled a secondary state-machine to actuate the arm, as shown in Fig.~\ref{fig:nested_smach}. The states, namely Home, Dig, Find Hauler, and Drop, had predefined configurations associated with them to simplify the manipulation motion planning problem. These predefined configurations are illustrated in Fig.~\ref{fig:ManipulationStates}.

During the Dig state, the first step was to lower the arm below the terrain and try to excavate the resource. However, the uncertainty on the Excavator's localization made the collection of a high percentage of the volatiles difficult, since this percentage was proportional to the distance between the bucket and the center of the volatile. To overcome this challenge, a search pattern was included for the arm bucket that increases the chances of finding the volatile. Once the volatile was found, the bucket continued to change the scoop direction by small-angle increments to increase the amount captured in each scoop and improve the quality of the localization updates. 

When any mass of volatile was detected in the bucket, the excavation state-machine transitioned to the Find Hauler state, which extended the arm in the direction of the Hauler's bin. The transition to the Drop state was only allowed when the Excavator got feedback that the Hauler had approached the Excavator so that the mass could be transferred from the bucket to the bin safely, when the Hauler was in the proper position. Then, the loop was repeated until all the mass of the volatile was collected.

During the excavation phase, if the Excavator's bucket successfully dug a volatile, it provided information about the rover position with respect to the map, given that the global positions of the volatiles were provided. Using the manipulator's forward kinematics and the current manipulator joint angles, the position of the end-effector in the global frame was estimated using the manipulator's forward kinematics formulation and the rover localization estimate. Then, this estimate was compared with the known location of the volatile, and the difference between them was used for state estimation as a pseudo-measurement update.

\subsection{Task 3 - CubeSat Localization and Rover Alignment}
\label{subsec:round3}

In this task, the first goal was to search for the CubeSat. To do this, we chose to toggle headlights to their high beam setting and tilted the camera up by 22.5~\si{deg} to visualize the CubeSat above the surface. Searching for the CubeSat used a combination of turn-in-place maneuvers and random driving. The rover started its mission by turning-in-place to see if it visualized the CubeSat. If the CubeSat was not visible, the rover's goal changed to find and approach the processing plant, which was always visible from the random starting location. Visual servoing with obstacle avoidance was used to approach the processing plant. Then, a random turn and a straight drive for some distance to the processing plant with the camera facing forward to avoid obstacles was followed by a turn-in-place maneuver with the camera looking upwards to find the CubeSat. The probability of longer driving distances increased as the mission time increased. If the CubeSat was not found, the rover returned to the processing plant and repeated the procedure. Once the CubeSat was found, the next goal was to move to a position that would improve the CubeSat position estimate. The rover tried to centralize the CubeSat in the camera image by turning and moving forward or backward. The CubeSat position was estimated using the methods defined in Section~\ref{subsec:cv}. The position of the CubeSat was reported to score points. 

After reporting the CubeSat position, the rover returned to the processing plant and reported within the region to score points. Then, the rover circulated the processing plant and aligned itself perpendicular to the fiducial marker using the stereo camera and planar LiDAR data to complete task~3.

\begin{figure}[t!]
    \centering
    \includegraphics[width=\linewidth]{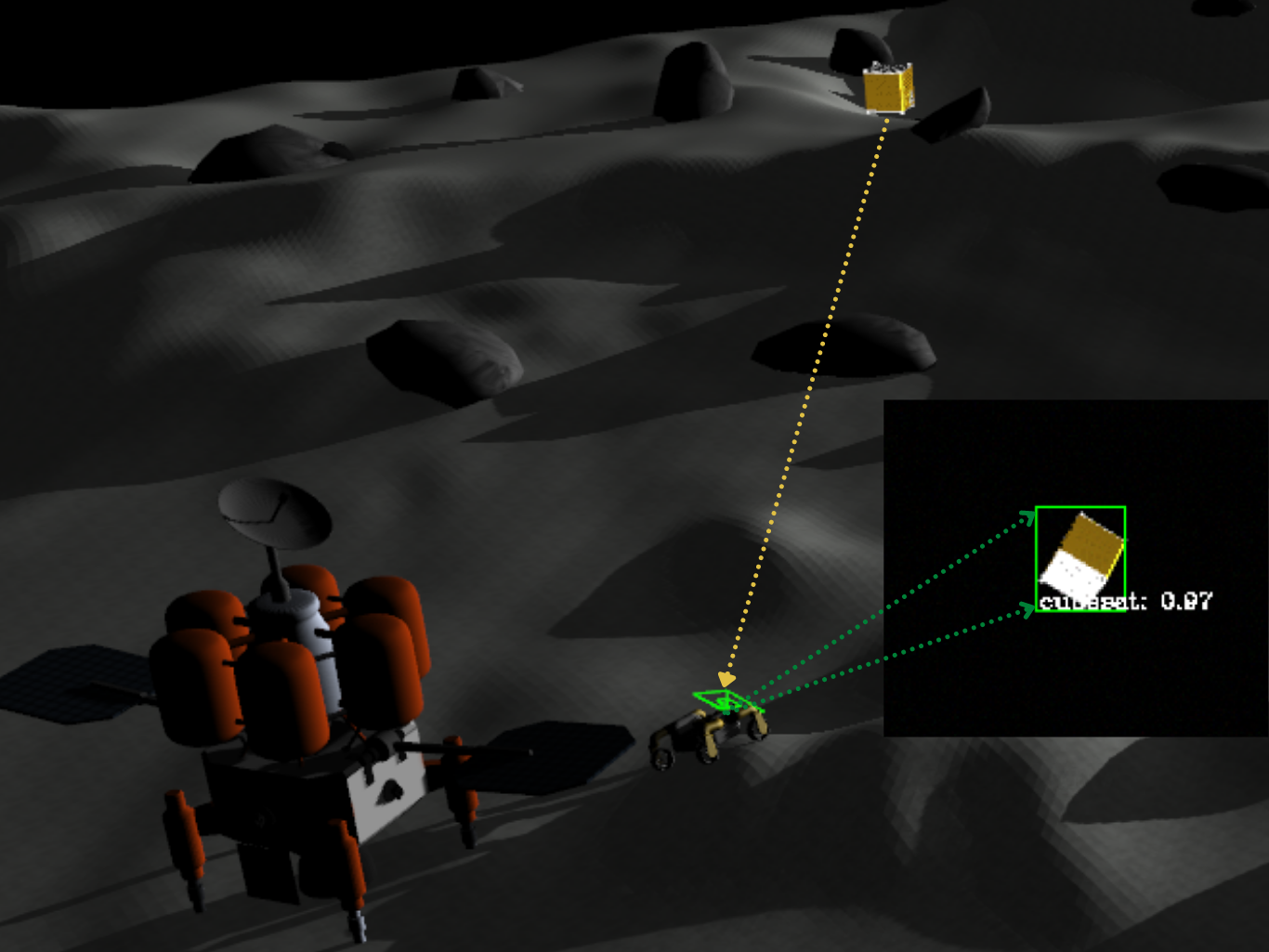}
    \caption{A depiction of CubeSat detection and position estimation in the simulation world. The rover centralized the CubeSat in the camera frame, then the average relative position was estimated.  }
    \label{fig:CubeSat}
\end{figure}

\subsubsection{CubeSat Localization}

Computer vision was used to detect the CubeSat, processing plant, and obstacles, to estimate their positions, and to align the rover with the processing plant's fiducial marker.  The approach presented in Section~\ref{subsec:cv} was used to accomplish these tasks.

\begin{figure}[htb!]
    \centering
    \includegraphics[scale=0.15]{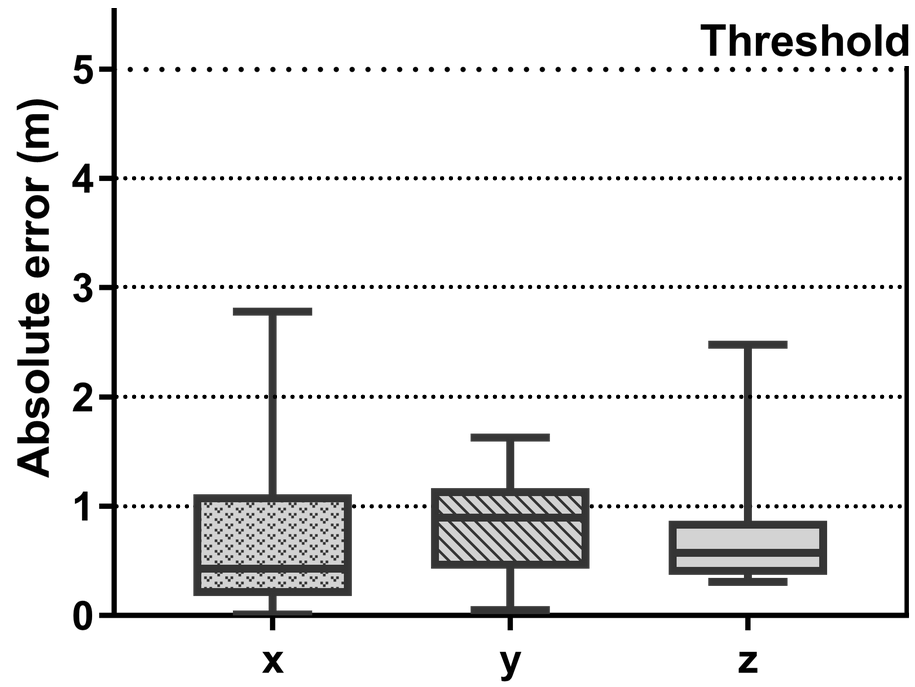}
    \caption{CubeSat localization: Comparison of the absolute error for the position coordinates with respect to the accuracy threshold adopted by the competition. The middle line in the boxes show the median absolute error value of 10 random simulation seeds. }
    \label{fig:CubeSat2}
\end{figure}

The average estimated position from 100 images with confidence above 90\% was used to estimate the CubeSat position. Figure~\ref{fig:CubeSat} demonstrates an example of the position estimation process as the rover had the CubeSat in the center of its camera image with 97\% confidence. A comparison between true pose and estimated pose was made for 10 runs in 10 randomly generated simulations. The error was calculated with reference to the world position where both the average and standard deviation error were less than 1\,\si{m} in all the axis as shown in Table~\ref{table:cubesat_error}. The maximum, minimum, and median absolute errors for CubeSat position estimation are also demonstrated in Fig.~\ref{fig:CubeSat2}. Notice that the localization provided was considered accurate if the position coordinates are within $\pm 5$ meters of the CubeSat location in the simulation environment~\cite{SRC2Rules2020}. 
\begin{table}[htb!]
\centering
\caption{Accuracy of the CubeSat position estimation in the simulation} 
\begin{tabular}{ lccc } 
 \hline
  Absolute Error (m) & x & y  & z \\ 
    \hline
    \hline
Seed \#32099 & 0.69 &	0.55 &	0.56 \\
Seed \#32793 & 0.11 &	0.55 &	0.44\\
Seed \#33720	& 0.20 &	1.02 &	0.38 \\
Seed \#39717	& \textbf{2.78} &	1.15 &	\textbf{2.48} \\
Seed \#50820	& 1.76 &	0.05 &	0.59 \\
Seed \#48796 & \textbf{0.01} &	1.08 &	0.35 \\
Seed \#49695	& 0.49 &	\textbf{1.63} &	1.04 \\
Seed \#51351	& 0.31 &	1.21 &	\textbf{0.31} \\
Seed \#51669 & 0.89 &	\textbf{0.07} &	0.81 \\
Seed \#48821 & 0.37 &	0.77 &	0.69 \\
 \hline
 \hline
Mean Error  (m) &	0.76&	0.81&	0.76 \\
Standard Dev.  (m) 	& 0.87&	0.51&	0.64 \\
RMS Error  (m)  &	1.12&	0.94&	0.98 \\
Median Error  (m) &	0.43&	0.90&	0.57 \\
 \hline
\end{tabular}

\label{table:cubesat_error}
\end{table}

\begin{figure}[htb!]
    \centering
    \includegraphics[width=\linewidth]{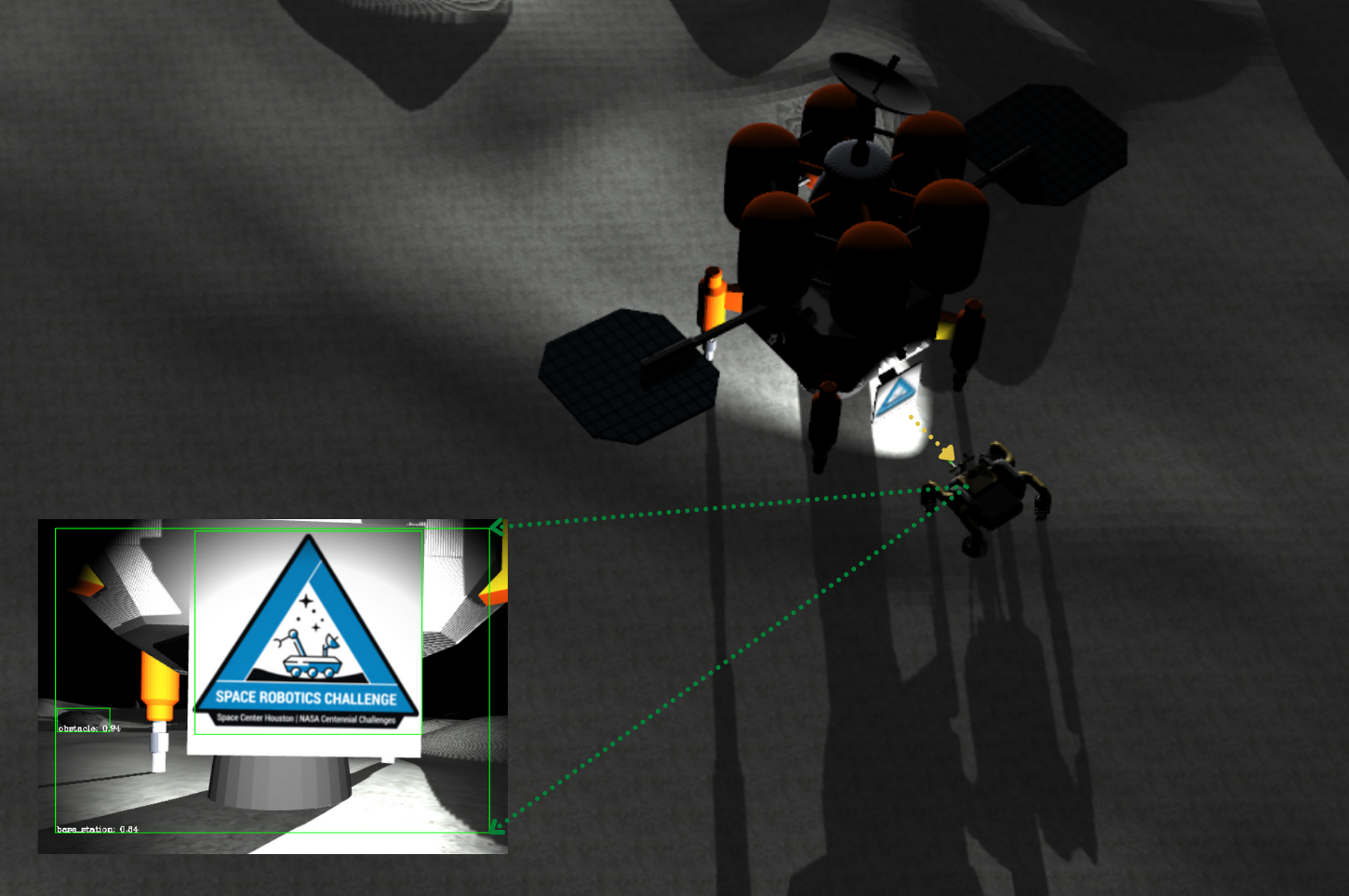}
    \caption{Rover alignment with the fiducial marker. This process was performed by iterating over centering the marker on the camera frame and aligning the rover with the base using the planar LiDAR. }
    \label{fig:alignment}
\end{figure}

\subsubsection{Alignment with the Processing Plant}

When the rover approaches the processing plant (base station), the 2D LiDAR was used to ensure the rover was within 3~\si{m} of the processing plant, and images were used to ensure the rover was facing the correct orientation. Crab-motion with the robot facing the processing plant was used for circling around the processing plant to align with the marker. The steering angles were set to obtain the radius needed to circulate it within the desired distance. 

A proportional (P) controller was used to control the steering angle and regulate the desired radius from the center of the processing plant. Once the fiducial marker was found on the image, a two-step proportional-integral (PI) control approach was used. First, the rover tried to center the fiducial marker in the image by turning-in-place (yaw). Then, the rover drove in the laterally to ensure that the left and right total distance from the laser was equivalent. Once both parameters were less than a threshold, based on the competition requirements, the rover reported that it was aligned with the processing plant. The rover alignment with the processing plant and the image streamed from the rover camera at that moment are shown in Fig.~\ref{fig:alignment}.

\section{Technical Challenges and Future Work}
\label{sec:lesson_learned}
In this section, the takeaways from the technical challenges that we faced during the qualification round are summarized. These challenges lead us to different research directions through the competition and help us significantly to qualify from the first round. Of the technical challenges faced in SRC2, a few proved to be more significant than the others and will require attention in the final round of the competition. First, obstacle avoidance was one of the most interesting and challenging problems we faced. Due to the undulating nature of the terrain, the dark background of space, and the obstacles having a similar texture to the terrain, identifying the obstacles accurately in the simulation environment was difficult. Also, the necessity of providing a real-time obstacle avoidance capability was challenging due to the limitations of the provided sensors. To circumvent the possible failures in the obstacle detection, the system ultimately relied upon several hand-tuned checks and behaviors, such as verifying if the rover got stuck in front of an obstacle or simply clearing the local motion planner's costmap before executing a navigation plan to avoid the accumulation of artifacts due to communication delay in point cloud registration. Advancing to the final round of the competition, more significant efforts on robust solutions to obstacle avoidance are foreseen as a continued challenge. Second, for lightweight, independent systems, the communication between the nodes using ROS framework is not a critical problem. However, to improve the autonomous capabilities of the rover, the communication between the nodes becomes more complex due to their inter-dependency. To alleviate this problem, specific ROS services dedicated to individual procedures were used, such as initialization maneuvers, immobility recoveries, and braking services. Moving forward, efforts on more efficiently sharing data (e.g., perceived objects, sensor data) will become a challenge as the number of robots increase.

The competition had three tasks with specific challenges, as mentioned before. It was beneficial to focus early development on core capabilities and be flexible on high-level strategies to overcome those challenges. For example, the computer vision detector was a core capability developed for the CubeSat detection. However, once this capability became available, we were able to update our strategy to leverage it for homing and obstacle avoidance. Similarly, a 4WS driving controller capability was developed for resource localization (task 1) to reduce drifting and then it was used for robot alignment for other tasks. This became an essential key to our success.

For the final round of the competition, the challenges from all three tasks discussed previously will be combined in a single mission with up to six rovers being used simultaneously. The final round will then include a new set of capabilities such as coordinating a bigger team of autonomous robots which will interact with each other, navigating on and exploring unknown lunar terrain, and excavating, collecting, and transporting the resources to a processing plant given little prior knowledge of the volatile locations and significant mission constraints (e.g., time, energy, number of robots).

\section{Conclusion}
\label{sec:conclusion}
This paper provided an overview of our solution for the Space Robotics Challenge Phase 2 qualification round, which required developing a cooperative autonomous robotic system. To share our experiences and insights gained through participation in a NASA challenge competition with the community, we presented the specific capabilities implemented by our team to support autonomous resource localization, resource excavation, and object detection tasks, along with a discussion of some of design trades we faced and an analysis of the performances obtained on some of the more challenging aspects of the competition . In the end, our submission was amongst the top 6 teams that secured the maximum qualification round prize heading into the finals.

\section*{Acknowledgment}
The authors would like to thank Nicholas Ohi, Chizhao Yang, Matteo de Petrillo, Rogerio Lima and Trevor Smith for their contributions on the qualification round of the competition, and Ali Baheri for helping review this paper.
Team Mountaineers would like to thank Benjamin M. Statler College of Engineering and Mineral Resources at West Virginia University for sponsoring our team in the Space Robotics Challenge Phase 2.

\bibliographystyle{IEEEtran}
\bibliography{references} 

\end{document}